\documentclass[10pt]{article} % For LaTeX2e
\usepackage[accepted]{tmlr}
\usepackage{url}
\usepackage{microtype}
\usepackage{graphicx,wrapfig,lipsum}
\usepackage{booktabs} % for professional tables
\usepackage{amsmath}
\usepackage{amsfonts}
\usepackage{multirow}
\usepackage{caption}
\usepackage{subcaption}
\usepackage{accents}
\usepackage{ntheorem}
\theoremseparator{:}
\newtheorem{hyp}{Assumption}
\usepackage{wrapfig}
\usepackage{tikz}
\usetikzlibrary{positioning,shapes,arrows}
\usepackage[ruled]{algorithm}
\usepackage{algpseudocode} 
\algnewcommand\algorithmicforeach{\textbf{for each}}
\algdef{S}[FOR]{ForEach}[1]{\algorithmicforeach\ #1\ \algorithmicdo}

\usepackage{mathtools,nccmath}

\newcommand\figref{Figure~\ref}
\newcommand\tabref{Table~\ref}
\newcommand\appref{Appendix~\ref}
\newcommand\hypref{A\ref}
\newcommand\secref{Section~\ref}

\title{Robust Hybrid Learning With Expert Augmentation}

\author{\name Antoine Wehenkel \email awehenkel@apple.com \\
  \addr Apple
  \AND
  \name Jens Behrmann \email j\_behrmann@apple.com\\
  \addr Apple 
  \AND
  \name Hsiang Hsu \email hsianghsu@g.harvard.edu\\
  \addr Harvard 
  \AND
  \name Guillermo Sapiro \email gsapiro@apple.com\\
  \addr Apple 
  \AND
  \name Gilles Louppe \email g.louppe@uliege.be\\
  \addr University of Li{\`e}ge 
  \AND
  \name Jörn-Henrik Jacobsen \email jhjacobsen@apple.com\\
  \addr Apple  }

\def\month{02}  % Insert correct month for camera-ready version
\def\year{2023} % Insert correct year for camera-ready version
\def\openreview{\url{https://openreview.net/forum?id=oe4dl4MCGY}} % Insert correct link to OpenReview for camera-ready version

\begin{document}

\maketitle

\begin{abstract}
Hybrid modelling reduces the misspecification of expert models by combining them with machine learning (ML) components learned from data. Similarly to many ML algorithms, hybrid model performance guarantees are limited to the training distribution. Leveraging the insight that the expert model is usually valid even outside the training domain, we overcome this limitation by introducing a hybrid data augmentation strategy termed \textit{expert augmentation}. Based on a probabilistic formalization of hybrid modelling, we demonstrate that expert augmentation, which can be incorporated into existing hybrid systems, improves generalization. We empirically validate the expert augmentation on three controlled experiments modelling dynamical systems with ordinary and partial differential equations. Finally, we assess the potential real-world applicability of expert augmentation on a dataset of a real double pendulum.

\begin{figure}[ht!]
\centering
\includegraphics[width=.6\textwidth]{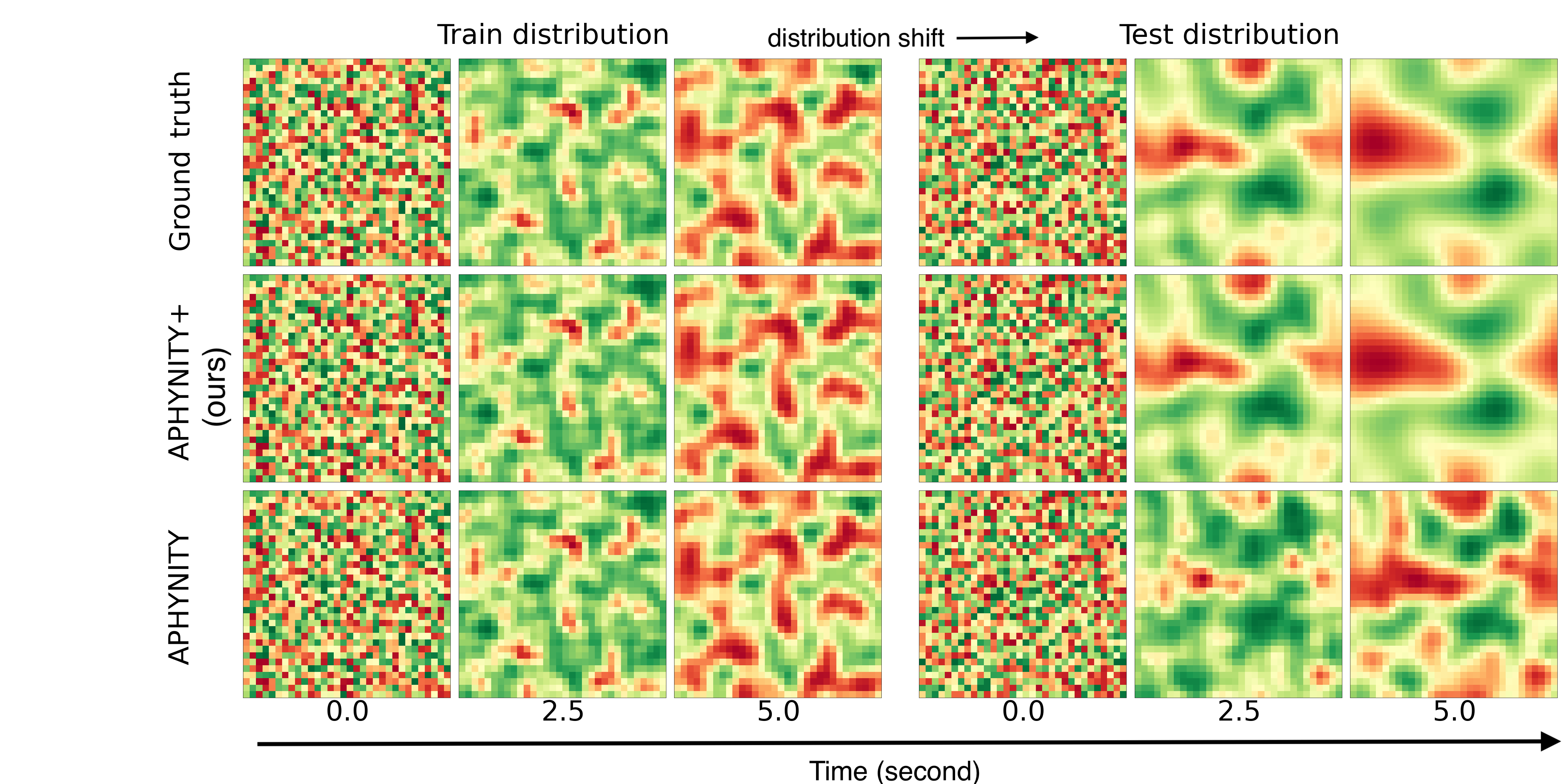}
\vspace{-.75em}

\caption{\small APHYNITY, an existing hybrid modelling strategy, is unable to predict accurately the dynamic of a 2D diffusion reaction for a shifted test distribution, although it predicts well configurations that follow the training distribution. APHYNITY+, the same model fine-tuned with our expert augmentation, generalizes to shifted distributions as expected from the validity of the underlying physics.}\label{fig:diffusion_shift}

\end{figure} 
\vspace{-2em}
\end{abstract}

%%%%%%%%%%%%%%%%%%%%%%%%%%%%%%%%%% 1 INTRODUCTION %%%%%%%%%%%%%%%%%%%%%%%%%%%%%%%%%%
\section{Introduction}

% \begin{wrapfigure}{r}{.51\textwidth}
% \vspace{-1em}
% \centering{
% \caption{APHYNITY, an existing hybrid modelling strategy, is unable to predict accurately the dynamic of a 2D diffusion reaction for a shifted test distribution although it predicts well configuration that follows the training distribution. On the opposite, APHYNITY+, the same model fine-tuned with our data augmentation, generalizes to shifted distributions as expected from the validity of the underlying physics.}\label{fig:diffusion_shift}
% \vspace{1em}
% \includegraphics[width=.5\textwidth]{figures/improved_diffusion.png}
% }
% \end{wrapfigure} 
Generalizing to unseen data is crucial to make a model applicable in the real world. When training and test data are independently and identically distributed~(IID), we assess the model generalization on a held-out subset of the training data. Unfortunately, the training and test scenarios do not entirely overlap in practice. This observation has motivated many recent research efforts to focus on the robustness of ML models  \citep{gulrajani2020search,geirhos2020shortcut,koh2021wilds}. Common strategies can be broadly grouped into two categories. The first class of methods aims to align properties of the model (e.g., invariance, equivariance or monotonicity) with expertise in the problem of interest \citep{cubuk2019autoaugment, graph_net_molecule, universal_equivariance, alphago}. The second category is data-focused \citep{groupDRO_ICLR, IRM, krueger2021out, creager2021environment}, and leverages variations present in the training data, e.g., some methods minimize the worst-case sub-group performance, to achieve robustness.

The data-oriented methods, which include Group-DRO~\citep{groupDRO_ICLR} and Invariant Risk Minimization~\citep[][IRM]{IRM}, present the advantage to specify the invariances implicitly via domains or environments. However, these methods rely on variations in the training data, which may be insufficient when the problem is too complex, or the variations of interest are absent from the training set. On the other hand, methods based on domain-specific expertise do not suffer from such limitations. Embedding expertise into a model can be done via architectural inductive biases \citep{lecun_cnn, gnn}, data augmentation \citep{cubuk2019autoaugment}, or a learning objective that enforces established symmetries of the problem~\citep{lagrangian_nn}. For example, simple data augmentation techniques combined with convolutions lead to excellent performance on natural image problems \citep{cubuk2019autoaugment}. Another natural approach to embedding expertise in ML models, and the one studied in this paper, is called hybrid learning, or sometimes gray-box modelling~\citep{willard2020integrating}. This framework combines an expert model (e.g., from first principle physics) with a learned component that improves the expert model so that the combination better fits real-world data. In hybrid learning, the expert model plays a central role and is supposed to provide a simple and well-grounded (parametric) description of the process considered. 
% The expert model is often motivated by the underlying physics system's. Hence, we will use the terms \textit{expert} model and \textit{physical} model interchangeably.

In recent works \citep{APHYNITY, HVAE, latent_ode_pharma, hl_1, hl_2, hl_3}, hybrid learning demonstrated success in complementing partial physical models and improving the inference of the corresponding parameters. We observe that current hybrid learning algorithms are sub-optimal in the amortized inference setting -- when we aim to build hybrid models that are valid for various test configurations. Contrary to the common belief that hybrid learning achieves better generalization than black box ML models, we argue and demonstrate that hybrid learning algorithms do not yet meet their promise regarding robustness in amortized settings. Although hybrid learning achieves strong performance on IID test distributions by exploiting the inductive bias of the expert models, their performance collapses when the test domain is not included in the training domain. This is unsatisfactory as the expert model is typically well-defined far outside the training distribution. 

% A test distribution not covered by the training data but for which an expert model exists often happens in the real world. For instance, \citet{latent_ode_pharma} apply hybrid learning to a pharmacological model describing the effect of a COVID-19 treatment for which only a limited quantity of real-world data is available. In this context, although the underlying biochemical dynamic of treatments is well modelled, data is often scarce and biased. Therefore, the hybrid model does not necessarily generalize to configurations that the pharmacological model well models if they are not part of the training set.

We introduce \textit{expert augmentations} for training augmented hybrid models~(AHMs), a procedure that extends the range of validity of hybrid models and improves generalization, as pictured by \figref{fig:diffusion_shift}. Our contribution is to first formalise the hybrid learning problem as: 1) Learning a probabilistic model partially defined by the expert model; 2) Performing inference over this probabilistic hybrid model. In this context, we show that hybrid learning is vulnerable to distribution shifts for which the expert model is well defined (see Figure \ref{fig:diffusion_shift}, bottom row). Motivated by our analysis, we propose to fine-tune the hybrid model on an expert-augmented dataset that includes distribution shifts (see results of augmentation in Figure \ref{fig:diffusion_shift}, middle row). These expert augmentations only rely on the hybrid model itself, leveraging that the expert model is also well-defined outside of the training distribution. Our experiments on various controlled problems demonstrate that AHMs improve the generalization capabilities of state-of-the-art hybrid learning algorithms on synthetic and real-world data in the amortize setting.

%%%%%%%%%%%%%%%%%%%%%%%%%%%%%%%%%% 2 Hybrid learning %%%%%%%%%%%%%%%%%%%%%%%%%%%%%%%%%%
\section{Hybrid learning}
\begin{wrapfigure}{r}{0.5\linewidth}
\centering\centering{
\scalebox{1.1}{
    \begin{tikzpicture}[
          node distance=.5cm and .5cm,
          mynode/.style={draw,circle,text width=.6cm,align=center},
          known/.style={draw,circle,text width=.6cm,align=center, double distance = .5pt},
          phantom_node/.style={circle,text width=.6cm,align=center},
          halo/.style={line join=round,
          double,line cap=round,double distance=#1,shorten >=-#1/2,shorten <=-#1/2},
          halo/.default=7mm]
        \node[mynode] (zp) {$Z_e$};
        \node[known, right=of zp] (x) {$X$};
        \node[mynode, right=of x] (za) {$Z_a$};
        \node[mynode, below=of zp] (yp) {$Y_e$};
        \node[mynode, below=of za] (y) {$Y$};
        \path (zp) edge[-latex] (yp);
        \path (x) edge[-latex] (yp);
        \path (x) edge[-latex] (y);
        \path (yp) edge[-latex] (y);
        \path (za) edge[-latex] (y);
        \path (za) edge[-latex] (y);
        
        \coordinate (A) at ([yshift=.1cm,xshift=-.1cm]zp.north west);
        \coordinate (B) at ([yshift=.1cm,xshift=.15cm]x.north east);
        \coordinate (C) at ([xshift=-.1cm,yshift=-.15cm]yp.south west);
        \coordinate (D) at ([yshift=-1.1cm,xshift=.1cm]x.south);
        
        \coordinate (A_right) at ([yshift=.5cm,xshift=0cm]A);
        \node at (A_right) [right] {Expert model};

        \draw[dashed] plot[smooth cycle] coordinates {(A) (B) (D) (C)};

    \end{tikzpicture}
    }
    \caption{\small A hybrid probabilistic model which describes the relationship between a given input $X$ and the output $Y$ for a configuration of the system as defined by the latent variables $Z_e$ and $Z_a$. The prescribed expert model defines the conditional density $p(y_e|z_e, x)$, where $Y_e$ is an approximation of $Y$. Hybrid learning aims at learning the conditional distribution $p(y|z_a, y_e, x)$.}
    \label{fig:gen_bnet}
}
\end{wrapfigure}
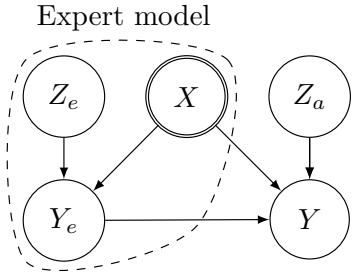

We formalize hybrid learning with the probabilistic model depicted in \figref{fig:gen_bnet}, and later rely on this formalism to show the benefits of the proposed proposed expert augmentation.
% In order to show that our proposed expert augmentations lead to robust models, we first formalize hybrid learning with the probabilistic model depicted in \figref{fig:gen_bnet}. 
In this Bayesian network, capital letters denote random variables (e.g., $Y$) and, we use calligraphic letters for the domain of the corresponding realization (e.g., $y \in \mathcal{Y}$). In our formalism, the expert model is a conditional density $p(y_e|x, z_e)$ that describes the distribution of the $expert$ response $Y_e$ to an input $x$ together with a parametric description of the system $z_e$, denoting expert parameters. We augment the expert model with the \textit{interaction model} which is a conditional distribution $p(y|x, y_e, z_a)$ that describes the distribution of the observation $Y$ given the input $x$, the expert model response $y_e$, and a parametric description of the interaction model $z_a$.

Our goal is to create a robust predictive model $p(y|x, (x_o, y_o))$ of the random variable $Y$, given the input $x$ together with independent observations $(x_o, y_o)$ of the same system, where the subscript $o$ denotes an observed quantity. As a concrete example, we consider predicting the evolution of a damped pendulum (described in \secref{sec:problems_description}) given its initial angle and speed ($x = \left[ \theta, \dot\theta\right]$) and a sequence of observations of the same pendulum. The expert model we assume is able to describe a frictionless pendulum whose dynamic is only characterized by one parameter $z_e := \omega_0$, denoting its fundamental frequency. The expert model is misspecified; it does not model the friction with a second parameter $z_a := \alpha$, the damping factor. In this problem, $(x_o, y_o)$ and $(x, y)$ are IID realizations of the same pendulum which corresponds, in general terms, to samples from $p(x, y|z_a, z_e)$ for some fixed but unknown values of $z_a$ and $z_e$. The expert variables $z_e$ (e.g., $\omega_0$) together with $z_a$ (e.g., $\alpha$) should accurately describe the system that produces $Y$ (e.g., the evolution of the pendulum's angle and speed along time) from $X$ (e.g., the initial pendulum's state). In our setting we assume that we are given a pair $(x_o, y_o)$ (e.g., past observations) from which we can accurately infer the state of the system $(z_a, z_e)$ as described by the interaction and expert models, and then predict the distribution of $Y$ for a given input $x$ (e.g., forecasting future observations) to the same system. Provided all probability distributions in \figref{fig:gen_bnet} are known, the Bayes optimal hybrid predictor $p_B$ is
\begin{align}
    p_B(y|x, (x_o, y_o)) = \mathbb{E}_{p(z_a, z_e|(x_o, y_o))}\left[ p(y|x, z_a, z_e) \right], \label{eq:hybrid_predictor}
\end{align}
as shown in \appref{app:proof_opt_bayes}.

% We observe that the Bayes optimal predictor explicitly depends on the posterior $p(z_a, z_e|(x_o, y_o))$ which is itself a function of the marginal distribution over $z_e$. This may preclude the existence of a good predictor that is invariant to shift of $p(z_e)$. 
In the amortized setting, we aim to learn a model of both the predictive model $p(y|x, z_a, z_e)$ and of the posterior over the parameters $p(z_a, z_e|(x_o, y_o))$. We will see that existing hybrid learning algorithms neglect the importance of building a robust encoder $p(z_a, z_e|(x_o, y_o))$ to make predictions in out-of-distribution (OOD) settings. In particular, we note that the interaction between $z_e$ and $y$ is essentially defined by the expert model. Thus it should be possible, and preferable, to learn a predictive model of $Y$ whose performance guarantees are as independent as possible from the training distribution of the expert variables $z_e$. Indeed, the validity of the expert model often go beyond the seen training examples. However, we demonstrate below that existing hybrid learning algorithms' performance collapses when the distribution of $z_e$ shifts. The expert augmentation introduced in this paper improves the robustness of hybrid models to such shifts.

% However, in the following we will consider that the pair $(x_o, y_o)$ contains enough information about the parameters $z_a, z_e$. As a consequence, the posterior distribution shrinks around the correct parameters value and the effect of the prior becomes negligible.

\subsection{Hybrid generative modelling}
As common in the hybrid learning literature, we consider expert models that are deterministic although our discussion clearly extends to stochastic expert models. The expert model describes the system as a function $f_e : \mathcal{X} \times \mathcal{Z}_e \rightarrow \mathcal{Y}_e$ that computes the response $y_e$ to an input $x$, parameterized by expert variables $z_e$. The goal of hybrid modelling is to augment the expert model with a component learned from data as depicted in \figref{fig:gen_bnet}. Formally, given a dataset $\mathcal{D} = \{(x^{(i)}, y^{(i)})\}_{i=1}^N$ of $N$ IID samples, we aim to learn the interaction model $p_\theta(y|x, y_e, z_a)$ that fits the data well but is close to the expert model. Defining closeness is hard and is an application-dependent modelling choice beyond the scope of this paper. However, common metrics include L2-distance between expert and hybrid outputs or Kullback-Leibler (KL) divergence between the marginal distributions of $Y$ and $Y_e$.

Learning a model that is close to the expert model and fits the training data well is a hard problem. However, the APHYNITY algorithm \citep{APHYNITY} and the Hybrid-VAE \citep[][HVAE]{HVAE} are two recent approaches that offer promising solutions to this problem. We now briefly describe how these methods approximate the Bayes optimal predictor of Equation~\eqref{eq:hybrid_predictor}. Our augmentation strategy is compatible (and effective) with both approaches.

\paragraph{APHYNITY.}
\citet{APHYNITY} consider hybrid learning for augmenting expert models defined as ordinary differential equation (ODE). They consider an additive hybrid model that should allow a perfect fitting of the data. In probabilistic terms, this assumption is equivalent to assuming the conditional distribution $p_\theta(y|x, y_e, z_a)$ is a Dirac distribution. Formally, they solve the following optimization problem
\begin{align}
    \min_{z_e, F_a} ||F_a|| \quad \text{s.t.} \quad \forall (x, y) \in \mathcal{D}, \forall t,& \frac{dy_t}{dt} = F_e(y_t, z_e) + F_a(y_t) \nonumber \\ &\text{with} \quad y_0 := x, \label{eq:APHYNITY}
\end{align}
where $||\cdot||$ is a norm operator on the function space, $F_a: \mathcal{Y}_t \rightarrow \mathcal{Y}_t$ is a learned function, $F_e: \mathcal{Y}_t \times \mathcal{Z}_e \rightarrow \mathcal{Y}_t$ defines the expert model and $\mathcal{D}$ is a dataset of initial states $x:=y_0$ and sequences $y \in \mathcal{Y} := (\mathcal{Y}_t)^k$, where $k$ is the number of observed timesteps. APHYNITY solves this problem with Lagrangian optimization and Neural ODEs~\citep{NODE} to compute derivatives. In the context of ODEs, the random variable $X$ is the initial state of the system at $t_0$ and $Y$ is the observed sequence of $k$ states between $t_0$ and $t_1$.

This formulation only considers learning a missing dynamic for one realization of the system described by \figref{fig:gen_bnet}, for a single $z_e$ (and $z_a$). However, we are interested in learning a hybrid model that works for the full set of systems described by \figref{fig:gen_bnet}. As suggested in \citet{APHYNITY}, we use an encoder network $g_\psi(\cdot, \cdot): \mathcal{X} \times \mathcal{Y} \rightarrow \mathcal{Z}_a \times \mathcal{Z}_e$ that corresponds to a Dirac distribution located at $g_\psi$ as the approximate posterior $q_\psi(z_a, z_e|x, y)$. The interaction model is a product of Dirac distributions whose locations correspond to the solution of the ODE 
\begin{align}
    \frac{dy_t}{dt} = F_e(y_t, z_e) + F_a(y_t, z_a; \theta), \quad y_0:=x.
    \label{eq:APH_ae}
\end{align} 
Hence the corresponding approximate Bayes predictor replaces the parameters $(z_a, z_e)$ in Equation~\eqref{eq:APH_ae} with the prediction of $g_\psi$ and predicts a product of Dirac distributions.

\paragraph{Hybrid-VAE (HVAE).}
In contrast to APHYNITY, the hybrid-VAE proposed by \citet{HVAE} is not limited to additive interactions between the expert model and the ML model, nor to ODEs. Instead, their goal is to learn the generative model described by \figref{fig:gen_bnet}. They achieve this with a variational auto-encoder (VAE) where the decoder specifically follows \figref{fig:gen_bnet}. Similarly to the amortized APHYNITY model, the encoder $g_\psi(x, y)$ predicts a posterior distribution over $z_a$ and $z_e$, and the model is trained with the classical Evidence Lower Bound on the likelihood~(ELBO). \citet{HVAE} observe that relying only on an architectural inductive bias and maximum likelihood training is not enough to ground the generative model to the expert equations. They propose to add three regularizers $R_{PPC}, R_{DA, 1},\text{ and } R_{DA, 2}$ that encourage the generative model to rely on the expert model. The final objective is 
\begin{align}
    \max_{\theta, \psi}\, &\mathbb{E}_{\mathcal{D}}\left[ \text{ELBO} ((x, y); \psi, \theta)\right] + \alpha R_{PPC} + \beta R_{DA, 1} + \gamma R_{DA, 2}. \label{eq:hvae_loss}
\end{align}
The first regularizer, $R_{PPC}$, encourages the marginal distribution of samples generated by the complete model to be close to the marginal distribution that would be only generated by the physical model. The two other regularizers specifically require the encoder network for $z_e$ to be made of two sub-networks. The first network filters the observations to keep only what can be generated by the expert model alone; it predicts $z_a$ and remove its effect from $y$ to predict a filtered version of it. The second network maps the filtered observations to the posterior distribution over $z_e$. $R_{DA, 1}$ penalizes the objective if the observations generated by the expert model are not close to the filtered observations. Finally, $R_{DA, 2}$ relies on data augmentation with the expert model to enforce that the second sub-network correctly identifies the expert variables $z_e$ if the observations were correctly filtered by the first encoder. We refer the reader to \appref{app:HVAE} and \citet{HVAE} for more details on HVAE. For HVAE, the approximate predictor takes the form described by Equation~\eqref{eq:hybrid_predictor} where $p(z_a, z_e|(x_o, y_o))$ is approximated by the encoder $q_\psi(z_a, z_e|x, y)$ and $p(y|x, z_a, z_e)$ by the learned hybrid generative model.

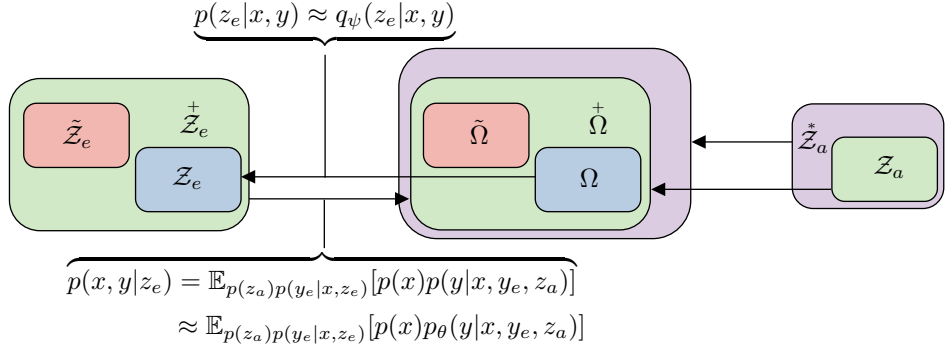
\begin{figure*}[h]
    \centering
    \tikzset{every picture/.style={line width=0.5pt}} %set default line width to 0.75pt        

\begin{tikzpicture}[x=0.75pt,y=0.75pt,yscale=-1,xscale=1,scale=0.8]

%All ALl omega
\draw  [color={rgb, 255:red, 0; green, 0; blue, 0 }  ,draw opacity=1 ][fill={rgb, 255:red, 219; green, 204; blue, 226 }  ,fill opacity=1. ] (247.5,65) .. controls (247.5,51.75) and (258.25,41) .. (271.5,41) -- (407.5,41) .. controls (420.75,41) and (431.5,51.75) .. (431.5,65) -- (431.5,137) .. controls (431.5,150.25) and (420.75,161) .. (407.5,161) -- (271.5,161) .. controls (258.25,161) and (247.5,150.25) .. (247.5,137) -- cycle ;

%All ze
\draw  [fill={rgb, 255:red, 210; green, 233; blue, 200 }  ,fill opacity=1. ] (2,79.2) .. controls (2,68.6) and (10.6,60) .. (21.2,60) -- (133.8,60) .. controls (144.4,60) and (153,68.6) .. (153,79.2) -- (153,136.8) .. controls (153,147.4) and (144.4,156) .. (133.8,156) -- (21.2,156) .. controls (10.6,156) and (2,147.4) .. (2,136.8) -- cycle ;
%all omega
\draw  [fill={rgb, 255:red, 210; green, 233; blue, 200 }  ,fill opacity=1. ] (255,78.7) .. controls (255,68.1) and (263.6,59.5) .. (274.2,59.5) -- (386.8,59.5) .. controls (397.4,59.5) and (406,68.1) .. (406,78.7) -- (406,136.3) .. controls (406,146.9) and (397.4,155.5) .. (386.8,155.5) -- (274.2,155.5) .. controls (263.6,155.5) and (255,146.9) .. (255,136.3) -- cycle ;

%Rounded Rect [id:dp6639592793786497] 
\draw [fill={rgb, 255:red, 241; green, 183; blue, 176 }  ,fill opacity=1. ]  (11,83.7) .. controls (11,79.23) and (14.63,75.6) .. (19.1,75.6) -- (69.4,75.6) .. controls (73.87,75.6) and (77.5,79.23) .. (77.5,83.7) -- (77.5,108) .. controls (77.5,112.47) and (73.87,116.1) .. (69.4,116.1) -- (19.1,116.1) .. controls (14.63,116.1) and (11,112.47) .. (11,108) -- cycle ;
%Rounded Rect [id:dp6448074412564562] 
\draw [fill={rgb, 255:red, 184; green, 205; blue, 225 }  ,fill opacity=1. ]  (81.5,111.6) .. controls (81.5,107.13) and (85.13,103.5) .. (89.6,103.5) -- (139.9,103.5) .. controls (144.37,103.5) and (148,107.13) .. (148,111.6) -- (148,135.9) .. controls (148,140.37) and (144.37,144) .. (139.9,144) -- (89.6,144) .. controls (85.13,144) and (81.5,140.37) .. (81.5,135.9) -- cycle ;

%Rounded Rect [id:dp7160429142184185] 
\draw  [fill={rgb, 255:red, 241; green, 183; blue, 176 }  ,fill opacity=1. ]  (263,83.7) .. controls (263,79.23) and (266.63,75.6) .. (271.1,75.6) -- (321.4,75.6) .. controls (325.87,75.6) and (329.5,79.23) .. (329.5,83.7) -- (329.5,108) .. controls (329.5,112.47) and (325.87,116.1) .. (321.4,116.1) -- (271.1,116.1) .. controls (266.63,116.1) and (263,112.47) .. (263,108) -- cycle ;
%Rounded Rect [id:dp8333286965822135] 
\draw [fill={rgb, 255:red, 184; green, 205; blue, 225 }  ,fill opacity=1. ]  (333.5,111.6) .. controls (333.5,107.13) and (337.13,103.5) .. (341.6,103.5) -- (391.9,103.5) .. controls (396.37,103.5) and (400,107.13) .. (400,111.6) -- (400,135.9) .. controls (400,140.37) and (396.37,144) .. (391.9,144) -- (341.6,144) .. controls (337.13,144) and (333.5,140.37) .. (333.5,135.9) -- cycle ;

\draw  [color={rgb, 255:red, 0; green, 0; blue, 0 }  ,draw opacity=1 ][fill={rgb, 255:red, 184; green, 233; blue, 134 }  ,fill opacity=0.3 ][fill={rgb, 255:red, 219; green, 204; blue, 226 }  ,fill opacity=1. ]  (495.5,89.2) .. controls (495.5,81.91) and (501.41,76) .. (508.7,76) -- (577.8,76) .. controls (585.09,76) and (591,81.91) .. (591,89.2) -- (591,128.8) .. controls (591,136.09) and (585.09,142) .. (577.8,142) -- (508.7,142) .. controls (501.41,142) and (495.5,136.09) .. (495.5,128.8) -- cycle ;
%Rounded Rect [id:dp6378499267772109] 
\draw  [color={rgb, 255:red, 0; green, 0; blue, 0 }  ,draw opacity=1 ] [fill={rgb, 255:red, 210; green, 233; blue, 200 }  ,fill opacity=1. ](520.5,105.1) .. controls (520.5,100.63) and (524.13,97) .. (528.6,97) -- (578.9,97) .. controls (583.37,97) and (587,100.63) .. (587,105.1) -- (587,129.4) .. controls (587,133.87) and (583.37,137.5) .. (578.9,137.5) -- (528.6,137.5) .. controls (524.13,137.5) and (520.5,133.87) .. (520.5,129.4) -- cycle ;

%Straight Lines [id:da18470078580218108] 
\draw    (149.5,122) -- (333.5,122) ;
\draw [shift={(148,122)}, rotate = 0] [fill={rgb, 255:red, 0; green, 0; blue, 0 }  ][line width=0.28]  [draw opacity=0] (8.93,-4.29) -- (0,0) -- (8.93,4.29) -- cycle    ;
%Straight Lines [id:da9950537909606885] 
\draw    (153,135.9) -- (253,135.9) ;
\draw [shift={(254,135.9)}, rotate = 180] [fill={rgb, 255:red, 0; green, 0; blue, 0 }  ][line width=0.08]  [draw opacity=0] (8.93,-4.29) -- (0,0) -- (8.93,4.29) -- cycle    ;
%Straight Lines [id:da27925255270157745] 
\draw    (201,45) -- (201,122.5) ;
%Straight Lines [id:da7038846140290534] 
\draw    (200.5,136.5) -- (200.5,167) ;
%Straight Lines [id:da3444837910426306] 
\draw    (409.5,129.9) -- (520,129.9) ;
\draw [shift={(406.5,129.9)}, rotate = 0] [fill={rgb, 255:red, 0; green, 0; blue, 0 }  ][line width=0.08]  [draw opacity=0] (8.93,-4.29) -- (0,0) -- (8.93,4.29) -- cycle    ;

\draw    (434,100) -- (495.5,100) ;
\draw [shift={(432,100)}, rotate = 0] [fill={rgb, 255:red, 0; green, 0; blue, 0 }  ][line width=0.08]  [draw opacity=0] (8.93,-4.29) -- (0,0) -- (8.93,4.29) -- cycle    ;

% Text Node
\draw (103,114.4) node [anchor=north west][inner sep=0.75pt]    {$\mathcal{Z}_{e}$};
% Text Node
\draw (33.5,85) node [anchor=north west][inner sep=0.75pt]    {$\tilde{\mathcal{Z}}_{e}$};
% Text Node
\draw (107.5,72.9) node [anchor=north west][inner sep=0.75pt]    {$\accentset{+}{\mathcal{Z}}_e$};
% Text Node
\draw (365.5,73.4) node [anchor=north west][inner sep=0.75pt]    {$\accentset{+}{\Omega}$};
% Text Node
\draw (290.5,84.9) node [anchor=north west][inner sep=0.75pt]    {$\tilde{\Omega}$};
% Text Node
\draw (360,115) node [anchor=north west][inner sep=0.75pt]    {$\Omega$};
% Text Node
\draw (545,107) node [anchor=north west][inner sep=0.75pt]    {$\mathcal{Z}_{a}$};
% Text Node
\draw (497,86) node [anchor=north west][inner sep=0.75pt]    {$\accentset{\ast}{\mathcal{Z}}_a$};
% Text Node
\draw (117,10) node [anchor=north west][inner sep=0.75pt]    {$\underbrace{p( z_{e} |x,y) \approx q_{\psi }( z_{e} |x,y)}$};
% Text Node
\draw (37,167.9) node [anchor=north west][inner sep=0.75pt]    {$\overbrace{p( x,y|z_{e}) =\mathbb{E}_{p( z_{a}) p( y_{e} |x, z_{e} )}[ p( x) p( y|x,y_{e}, z_{a})]}$};
% Text Node
\draw (106.5,206.9) node [anchor=north west][inner sep=0.75pt]    {$\approx \mathbb{E}_{p( z_{a}) p( y_{e} |x, z_{e})}[ p( x) p_{\theta }( y|x,y_{e}, z_{a})]$};

\end{tikzpicture}
    \vspace{-.75em}
    \caption{\small Visualization of the distribution shifts considered in this work. The train support $\Omega$ of $(x, y)$ results from $(z_a, z_e) \in \mathcal{Z}_a \times \mathcal{Z}_e$. The test supports (in red) are denoted with a tilde symbols as $\tilde{\mathcal{Z}}_e$ for $z_e$ and $\tilde{\Omega}$ for $(x, y)$. The augmented support $\accentset{+}{\Omega}$ (in green) includes both train and test scenarios and corresponds to $(z_a, z_e) \in \mathcal{Z}_a \times \accentset{+}{\mathcal{Z}}_e$. The outer violet domain that includes $\accentset{+}{\Omega}$ depicts one of our experiment in which the domain of $z_a$ is also shifted. Hybrid modelling algorithms alone may learn a mapping $p_\theta: \accentset{+}{\mathcal{Z}}_e \rightarrow \accentset{+}{\Omega}$ but augmentation is necessary to learn the inverse mapping $q_\psi: \accentset{+}{\Omega} \rightarrow \accentset{+}{\mathcal{Z}}_e$.}
    \label{fig:ood_vizu}
    \vspace{-1em}
\end{figure*}

%%%%%%%%%%%%%%%%%%%%%%%%%%%%%%%%%% 3 Robust hybrid learning %%%%%%%%%%%%%%%%%%%%%%%%%%%%%%%%%%
\section{Robust hybrid learning}
We now formalize our definition of out of distribution~(OOD) and robustness. In general, a test scenario is OOD if the joint test distribution $\tilde{p}(x, y)$ is different from the training distribution $p(x, y)$, that is $d(\tilde{p}, p) > 0$ for any properly defined divergence or distance $d$ between distributions. In the following, we reduce our discussion to a  sub-class of distribution shifts for which the marginal train and test distributions over $z_e$ may be different, $d(p(z_e),  \tilde{p}(z_e)) > 0$, but the marginals of $z_a$ and $x$ are constant. As a consequence, the joint distribution of $(x, y)$ pairs is also shifted. Formally, the training and test distributions are respectively defined as
\begin{align}
    p(x, y) := \mathbb{E}_{p(z_e) p(z_a) p(y_e|x, z_e)}\left[p(x) p(y|x, y_e, z_a) \right], \nonumber\\
    \tilde{p}(x, y) := \mathbb{E}_{\tilde{p}(z_e) p(z_a) p(y_e|x, z_e)}\left[p(x) p(y|x, y_e, z_a) \right]. \nonumber
\end{align}
In this context, we demonstrate, theoretically and empirically, that classical hybrid models fail. To address this failure, we introduce \textit{augmented hybrid models} and show that, under some assumptions, they achieve optimal performance on both the train and test distributions. 

Our goal is to learn a predictive model 
\begin{align}
    p_{\theta, \psi}(y|x, (x_o, y_o)) = \mathbb{E}_{\substack{q_\psi(z_a, z_e|x_o, y_o) \\ p(y_e|x, z_e)}}\left[ p_\theta(y|x, y_e, z_a)\right] \nonumber
\end{align} 
that is \textit{exact} on both the train and test domains when they follow the  aforementioned training and testing distribution shifts. We say that a learned predictive model $\hat{p}(a|b)$ is $\mathcal{E}\textit{-exact}$, or \textit{exact} on the sample space $\mathcal{E}$, if $\hat{p}(a|b) = p(a|b) \quad \forall (a, b) \in \mathcal{E}$. Here we qualify a predictive model as \textit{robust} to a test scenario if its \textit{exactness} on the training domain is sufficient to ensure exactness on the test domain. 

We now define an augmented distribution $\accentset{+}{p}(z_e)$ over the expert variables whose support $\accentset{+}{\mathcal{Z}}_e$ includes the joint support $\mathcal{Z}_e \cup \tilde{\mathcal{Z}}_e$ between the train and test distribution of the physical parameters. As depicted in \figref{fig:ood_vizu}, we denote the corresponding support over the observation space $\mathcal{X} \times \mathcal{Y}$ as $\accentset{+}{\Omega}$ for the augmented distribution, $\Omega$ for the training distribution, and $\tilde{\Omega}$ for the test distribution. In this context, and with \textbf{\hypref{hyp:first}}, we may demonstrate that even under perfect learning, classical hybrid learning algorithms do not produce an $\tilde{\Omega}\textit{-exact}$ predictor while our augmentation strategy does.
\begin{hyp}[A\ref{hyp:first}] \label{hyp:first}
Hybrid modelling learns an interaction model $p_\theta(y|y_e, x, z_a)$ that is $\accentset{+}{\Omega}\textit{-exact}$. 
\end{hyp}
Although strong, \textbf{\hypref{hyp:first}} is consistent with the recent literature on hybrid modelling, which assumes that the expert model $p(y_e|x, z_e)$ is an accurate description of the system, thereby the interaction model $p_\theta(y|y_e, x, z_a)$ should not be overly complex. As an example, we consider an additive interaction model in our experiments for which extrapolation to unseen $y_e$ holds if the additive assumption is correct. That said, we still notice that the exactness of the interaction model $p_\theta$ on the augmented support $\accentset{+}{\Omega}$ is insufficient to prove that the predictive model $p_{\theta, \psi}$ is $\accentset{+}{\Omega}\textit{-exact}$. Indeed, the encoder $q_\psi$ is only trained on the training data and cannot rely on a strong inductive bias in contrast to $p_\theta$. Thus, even if the encoder is exact on the training distribution, the corresponding predictive model is not $\accentset{+}{\Omega}\textit{-exact}$. While the decoder's performance are not limited to the training scenarios thanks to the broader validity of the expert model, the encoder does not generalize to unseen settings as it is purely data-driven.

\begin{wrapfigure}{r}{0.5\linewidth}
\begin{minipage}{0.5\textwidth}
\begin{algorithm}[H]
\begin{algorithmic}[1]
	\caption{Expert augmented hybrid learning} \label{algo:augmentation}
        \State $\mathcal{D}:= \{(x^{(i)}, y^{(i)})\}_{i=1}^{N} \in (\mathcal{X} \times \mathcal{Y})^{N}$  \Comment{ {\footnotesize A training set}}
        \State $q_\psi(z_a, z_e | x, y)$ \Comment{ {\footnotesize A parametric encoder}}
        \State $p (y_e| x, z_e)$ \Comment{ {\footnotesize An expert model}}
        \State $p_\theta (y| x, y_e, z_a)$ \Comment{ {\footnotesize A parametric interaction model}}
        \State $l(x, y, \theta, \psi)$ \Comment{ {\footnotesize A hybrid learning objective function}}
        \State $p_{+}(z_e)$ \Comment{ {\footnotesize A prior distribution on $z_e$ that covers both train and test scenarios}}
    \Procedure{Training}{}      
    \State $\psi^\star, \theta^\star \gets \arg\min_{\psi, \theta} \mathbb{E}_{(x, y) \sim \mathcal{D}} \left[l(x, y, \theta, \psi) \right]$
    \State{{\footnotesize  $\theta^\star$ is frozen.}}
    \State $\accentset{+}{\mathcal{D}} \gets \Call{GenerateAugmentedSet}{}$
    \State $\mathcal{D} \gets \accentset{+}{\mathcal{D}} \cup \mathcal{D}$
    \State{\footnotesize Finetuning the encoder on the augmented training set:}
    \State\vspace*{-2em}
        \begin{fleqn}[1.4em]
        \setlength\belowdisplayskip{0pt}
        \begin{equation}
        \small
            \begin{multlined}[c]
            \psi^\star \gets \arg\min_{\psi} \mathbb{E}_{\mathcal{D}} \left[l(x, y, \theta^\star, \psi)\right]\\ - \mathbb{E}_{\accentset{+}{\mathcal{D}}}\left[\log q_\psi(z_e|x, y)\right] \nonumber
            \end{multlined}
        \end{equation}
        \end{fleqn}%    
        \State \Return $\psi^\star, \theta^\star$
    \EndProcedure
    \Procedure{GenerateAugmentedSet}{} 
    \State $\accentset{+}{\mathcal{D}} \gets \{\}$
    \ForEach{$(x_o, y_o) \in \mathcal{D}$}
        \State $z_a \sim q_{\psi^\star}(z_a, z_e | x_o, y_o) $
        \State $z_e \sim p_{+}(z_e)$
        \State $y_e \sim p(y_e| x, z_e)$
        \State $y \sim p_{\theta^\star} (y| x, y_e, z_a)$
        \State $\accentset{+}{\mathcal{D}} \gets \accentset{+}{\mathcal{D}} \cup \{(x_o, y, z_e)\}$
    \EndFor
    \State \Return $\accentset{+}{\mathcal{D}}$
        \EndProcedure

\end{algorithmic}
\end{algorithm}
\end{minipage}
\end{wrapfigure}

\subsection{Expert augmentation}
We propose a data augmentation strategy to improve the robustness of hybrid models to unseen test scenarios. Once trained, the hybrid model is composed of an encoder $q_\psi$ and an interaction model $p_\theta$ that are respectively $\Omega\textit{-}$ and $\accentset{+}{\Omega}\textit{-exact}$. We may create a new training distribution with a support over $\accentset{+}{\Omega}$ by sampling physical parameters $z_e$ from a distribution that covers $\accentset{+}{\mathcal{Z}}_e$. Then, we train the encoder $q_\psi$ on the augmented domain $\accentset{+}{\Omega}$, under perfect training the predictive model $p_{\theta, \psi}(y|x, (x_o, y_o))$ is $\accentset{+}{\Omega}\textit{-exact}$, hence exact on both train and test domains. The expert augmentation is formally described in Algorithm~\ref{algo:augmentation}, for further details see \appref{app:ahm_desc}.

Our learning strategy is grounded in existing hybrid modelling algorithms, and here, we focus on APHYNITY and HVAE. We first train an encoder $q_\psi$ and a decoder $p_\theta$ with a hybrid learning algorithm. Together with experts we then decide on a realistic distribution $\accentset{+}{p}(z_e)$ and create a new dataset $\accentset{+}{\mathcal{D}} := \{(z_e, x_i, y_i)\}^{\accentset{+}{N}}_{i=1}$ by sampling from the hybrid generative model defined by \figref{fig:gen_bnet} and the interaction model $p_\theta$. A notable difference between the augmented training set $\accentset{+}{\mathcal{D}}$ and the original training set $\mathcal{D}$ is that the former contains ground truth values for the expert's variables $z_e$. As we assume that the interaction model is $\accentset{+}{\Omega}\textit{-exact}$, we freeze it and only fine-tune the encoder $q_\psi$ on $\mathcal{D} \cup \accentset{+}{\mathcal{D}}$. We use a combination of the loss function $\ell$ of the original algorithm (e.g., Equation~\eqref{eq:hvae_loss} for HVAE, and the Lagrangian of Equation~\eqref{eq:APHYNITY} for APHYNITY) and a supervision on the latent variable objective to learn a decoder that~solves
\begin{align}
    \psi^\star &= \arg\min_{\psi} \mathbb{E}_{\accentset{+}{p}(z_e) p(z_a) p(x) p(y | x, z_e, z_a)}\left[ \ell(x, y; \theta, \psi) - \log q_\psi(z_e|x, y) \right], \label{eq:augm_loss}\\
    &\approx \arg\min_{\psi}  \frac{1}{N + \accentset{+}{N}}\sum_{(x, y) \in \mathcal{D} \cup \accentset{+}{\mathcal{D}}} \ell(x, y; \theta, \psi) - \frac{1}{\accentset{+}{N}}\sum_{(z_e, x, y) \in \accentset{+}{\mathcal{D}}}\log q_\psi(z_e|x, y). \nonumber
\end{align}
In our experiments we use a Gaussian distribution for the posterior, which is equivalent to a mean squared error~(MSE) loss on the physical parameters. We provide a detailed description of the expert augmentation scheme in \appref{app:ahm_desc}. In principle, the regularizers (the norm of $f_a$ for APHYNITY and $ R_{PPC}$, $R_{DA, 1}$, and $R_{DA, 2}$ for the Hybrid-VAE) are not necessary as Equation~\eqref{eq:augm_loss} only aims to improve the encoder whereas these terms aim to regularize the interaction model. However, in practice, we have observed that this does not matter and makes the implementation even more straightforward.

As a side note, we would like to emphasize the difference between the data augmentation proposed in this paper and the one from \citet{HVAE}. While HVAE also requires to sample new physical parameters $z_e$, it is only to ensure that a sub-part of the encoder is able to infer correctly $z_e$ given $y_e$. This augmentation does not contribute robustness to distribution shifts on $y$ in contrast to ours.

%%%%%%%%%%%%%%%%%%%%%%%%%%%%%%%%%% 4 Experiments %%%%%%%%%%%%%%%%%%%%%%%%%%%%%%%%%%
\section{Experiments}
We assess the benefits of expert augmentation on three synthetic problems and one real-world experiment that are described by the ODE
\begin{align}
    \frac{dy_{t}}{dt} = F_e(y_t; z_e) + F_a(y_t; z_a), \label{eq:generic_ode}
\end{align}
where $F_e: \mathcal{Y}_t \times \mathcal{Z}_e \rightarrow \mathcal{Y}_t$ is the expert model and $F_a: \mathcal{Y}_t \times \mathcal{Z}_a \rightarrow \mathcal{Y}_t$ complements it. In our notation $X$ is the initial state $y_0$ and the response $Y$ is the sequence of states $y_{1:t_1}:=[y_{i\Delta t}]_{i=1}^{t_1 \!/\! \Delta t}$. For all experiments we train the models to maximize $p_{\theta, \psi}(y=y_{1:t_1}|x=y_0)$ on the training data. We validate and test the models on the predictive distribution $p(y=y_{1:t_2}|x=y_0, x_o=y_0, y_o=y_{1:t_1})$, where $t_2 > t_1$ assesses the generalization over time. A brief description of the different problems is provided below. Readers interested to reproduce our experiments can check our code at \url{https://github.com/apple/ml-robust-expert-augmentations}.
\subsection{Synthetic experiments} \label{sec:problems_description}

\textbf{The damped pendulum} is often used as an example in the hybrid modelling literature \citep{APHYNITY, HVAE}. The system's state at time $t$ is $y_t = \begin{bmatrix}\theta_t & \dot \theta_t\end{bmatrix}^T$, where $\theta_t$ is the angle of the pendulum at time t and $\dot \theta_t$ its angular speed. The evolution of the state over time is described by Equation~\eqref{eq:generic_ode}, where $z_e:=\omega$, $z_a=\alpha$ and
\begin{align}
    F_e := \begin{bmatrix}\dot \theta_t &  -\omega_0^2 \sin \theta_t   \end{bmatrix}^T \quad \text{and} \quad F_a := \begin{bmatrix}0 &  -\alpha \dot \theta_t \end{bmatrix}^T.
\end{align}
The corresponding systems are defined by the damping factor $\alpha$ and $\omega_0$, the fundamental frequency of the pendulum.

\textbf{The RLC series circuits} are electrical circuits made of 3 electrical components that may model a large range of transfer functions. A schematic of such circuit is shown in \figref{fig:RLC}. These models are often used in biology (e.g., the Hodgkin-Huxley class of models~\citep{hh-model}, in photoplethysmography~\citep{ppg-model}) and in electrical engineering to model the dynamics of various systems. The system's state at time $t$ is $y_t = \begin{bmatrix}U_t & I_t\end{bmatrix}^T$, where $U_t$ is the voltage over the capacitor and $I_t$ the current in the circuit. The evolution of the state over time is described by Equation~\eqref{eq:generic_ode}, where $z_e:=\{L, C\}$, $z_a= \{R\}$ and

\begin{align}
    F_e := \begin{bmatrix}\frac{I_t}{C} \\ \frac{1}{L} (V_t - U_t) \end{bmatrix} \quad \text{and} \quad F_a := \begin{bmatrix}0 \\- \frac{R}{C}I_t\end{bmatrix}.
\end{align} The dynamics described by the RLC circuit is more diverse than for the pendulum and the system can be hard to identify. This system is characterised by the resistance $R$, capacitance $C$, and inductance $L$, provided $V_t$ is known.

\textbf{The 2D reaction diffusion} was used by \citet{APHYNITY} to assess the quality of APHYNITY. It is a 2D FitzHugh-Nagumo on a $32 \times 32$ grid. The system's state at time $t$ is a $2\times 32 \times 32$ array $y_t = \begin{bmatrix}u_t & v_t\end{bmatrix}^T$. The evolution of the state over time is described by Equation~\eqref{eq:generic_ode}, where $z_e:=\{a, b\}$, $z_a= \{k\}$ and 
\begin{align}
    F_e := \begin{bmatrix}a \Delta u_t \\ b \Delta v_t \end{bmatrix} \quad \text{and} \quad F_a := \begin{bmatrix}R_u(u_t, v_t;k) \\R_v(u_t, v_t)\end{bmatrix},
\end{align}
where $\Delta$ is the Laplace operator, the local reaction terms are $R_u(u, v;k) = u - u^3 - k -v$ and $R_v(u, v) = u - v$.
This model is interesting to study as it considers a state space for which neural architectures may have a real advantage compared to other ML models. In particular, convolutional neural networks are effective for processing signals with spacial and/or temporal correlation.

In these experiments we analyze the effect of our data augmentation strategy on APHYNITY and HVAE. All models explicitly use the assumption that the interaction model follows the structure of Equation~\eqref{eq:generic_ode}. For each problem the validation and test sets are respectively IID and OOD with respect to the training distribution. The best models are always selected based on validation performance, that is with samples from $\Omega$. We provide additional details on the different expert models, dataset creation, and neural networks architectures in \appref{app:exp_details}.

\subsection{A real world dataset - the double pendulum}
\begin{figure}
\begin{subfigure}[b]{0.23\textwidth}
    \includegraphics[width=1.\textwidth]{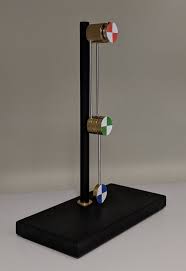}
    \caption{}\label{fig:pic_double_pendulum}
\end{subfigure}
\begin{subfigure}[b]{0.31\textwidth}
    \centering
\tikzset{every picture/.style={line width=0.75pt}} %set default line width to 0.75pt        

\begin{tikzpicture}[x=0.75pt,y=0.75pt,yscale=-1,xscale=1]

%Straight Lines [id:da020400525504394196] 
\draw [line width=1.5]    (300,40) -- (381,121) ;
%Straight Lines [id:da9803772546733409] 
\draw [line width=1.5]    (381,121) -- (336,212) ;
%Shape: Circle [id:dp231981160931096] 
\draw  [color={rgb, 255:red, 0; green, 0; blue, 0 }  ,draw opacity=1 ][fill={rgb, 255:red, 196; green, 53; blue, 53 }  ,fill opacity=1 ] (292,40) .. controls (292,35.58) and (295.58,32) .. (300,32) .. controls (304.42,32) and (308,35.58) .. (308,40) .. controls (308,44.42) and (304.42,48) .. (300,48) .. controls (295.58,48) and (292,44.42) .. (292,40) -- cycle ;
%Straight Lines [id:da7712629298677495] 
\draw [fill={rgb, 255:red, 230; green, 50; blue, 71 }  ,fill opacity=1 ] [dash pattern={on 4.5pt off 4.5pt}]  (300,40) -- (301,126) ;
%Straight Lines [id:da5175748156642541] 
\draw  [dash pattern={on 4.5pt off 4.5pt}]  (381,121) -- (382,207) ;
%Curve Lines [id:da5075474268744414] 
\draw    (348.19,195.41) .. controls (354.7,198.6) and (364.76,202.08) .. (381,202.64) ;
\draw [shift={(345.5,194)}, rotate = 28.93] [fill={rgb, 255:red, 0; green, 0; blue, 0 }  ][line width=0.08]  [draw opacity=0] (8.93,-4.29) -- (0,0) -- (8.93,4.29) -- cycle    ;
%Curve Lines [id:da12442521654035799] 
\draw    (300,121.64) .. controls (307.95,121.31) and (330.97,119.87) .. (356.07,99.6) ;
\draw [shift={(358,98)}, rotate = 139.85] [fill={rgb, 255:red, 0; green, 0; blue, 0 }  ][line width=0.08]  [draw opacity=0] (8.93,-4.29) -- (0,0) -- (8.93,4.29) -- cycle    ;
%Shape: Circle [id:dp8737737910353444] 
\draw  [color={rgb, 255:red, 0; green, 0; blue, 0 }  ,draw opacity=1 ][fill={rgb, 255:red, 105; green, 160; blue, 42 }  ,fill opacity=1 ] (373,121) .. controls (373,116.58) and (376.58,113) .. (381,113) .. controls (385.42,113) and (389,116.58) .. (389,121) .. controls (389,125.42) and (385.42,129) .. (381,129) .. controls (376.58,129) and (373,125.42) .. (373,121) -- cycle ;
%Shape: Circle [id:dp515687221781606] 
\draw  [color={rgb, 255:red, 0; green, 0; blue, 0 }  ,draw opacity=1 ][fill={rgb, 255:red, 196; green, 53; blue, 53 }  ,fill opacity=1 ] (292,40) .. controls (292,35.58) and (295.58,32) .. (300,32) .. controls (304.42,32) and (308,35.58) .. (308,40) .. controls (308,44.42) and (304.42,48) .. (300,48) .. controls (295.58,48) and (292,44.42) .. (292,40) -- cycle ;
%Shape: Circle [id:dp5455385959514643] 
\draw  [color={rgb, 255:red, 0; green, 0; blue, 0 }  ,draw opacity=1 ][fill={rgb, 255:red, 59; green, 120; blue, 192 }  ,fill opacity=1 ] (328,212) .. controls (328,207.58) and (331.58,204) .. (336,204) .. controls (340.42,204) and (344,207.58) .. (344,212) .. controls (344,216.42) and (340.42,220) .. (336,220) .. controls (331.58,220) and (328,216.42) .. (328,212) -- cycle ;

% Text Node
\draw (325.5,119) node [anchor=north west][inner sep=0.75pt]   [align=left] {$\displaystyle \theta _{1}$};
% Text Node
\draw (357.5,205) node [anchor=north west][inner sep=0.75pt]   [align=left] {$\displaystyle \theta _{2}$};
% Text Node
\draw (346,60.5) node [anchor=north west][inner sep=0.75pt]   [align=left] {$\displaystyle l_{1}$};
% Text Node
\draw (340.5,150) node [anchor=north west][inner sep=0.75pt]   [align=left] {$\displaystyle l_{2}$};
% Text Node
\draw (393.5,110.5) node [anchor=north west][inner sep=0.75pt]   [align=left] {$\displaystyle m_{1}$};
% Text Node
\draw (304.5,202) node [anchor=north west][inner sep=0.75pt]   [align=left] {$\displaystyle m_{2}$};
\end{tikzpicture}
\vspace{1em}
\caption{}\label{fig:double_pendulum_sketch}
\end{subfigure}
\begin{subfigure}[b]{0.46\textwidth}
    \includegraphics[width=1.\textwidth]{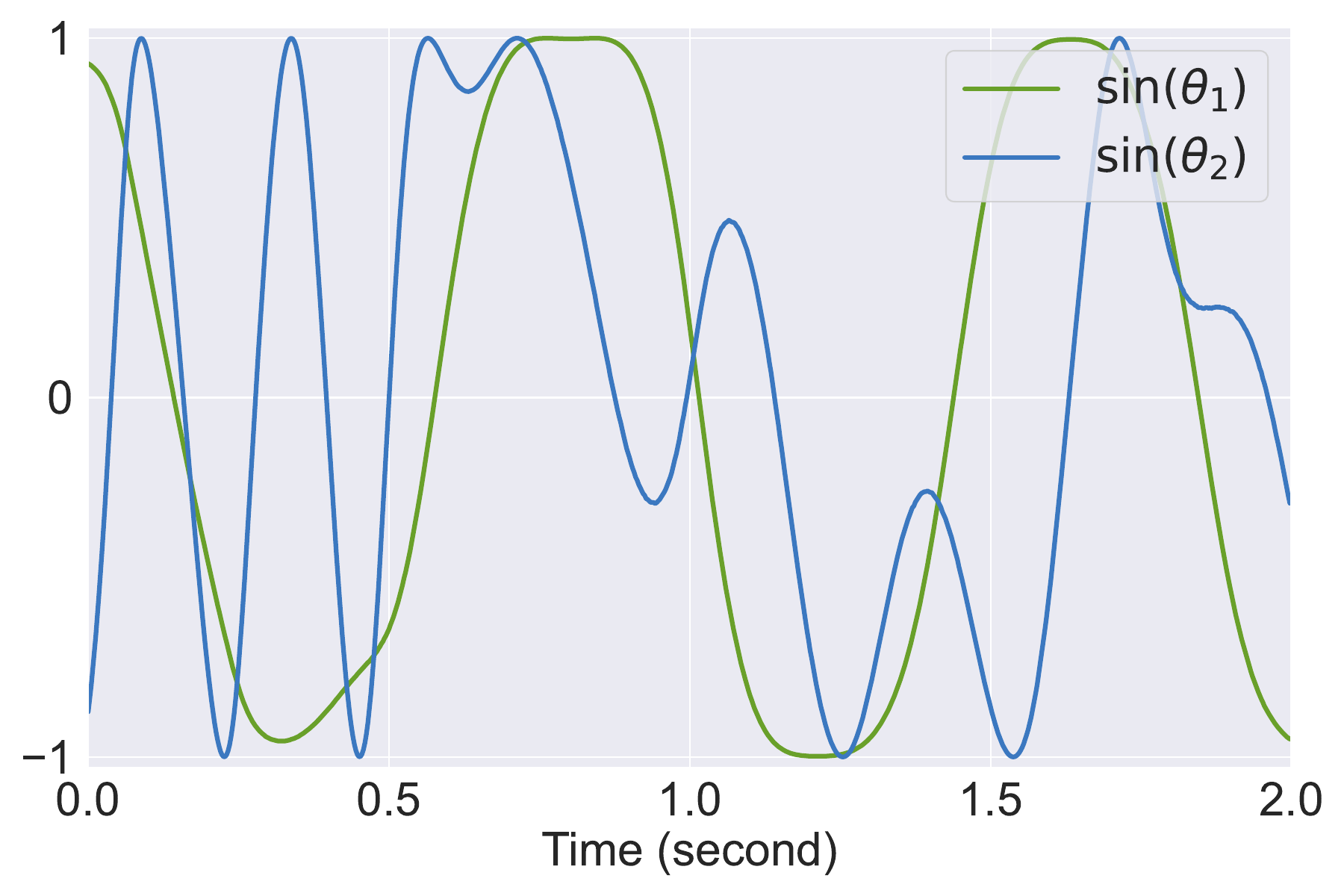}
    \caption{}\label{fig:double_pendulum_processed}
\end{subfigure}
\caption{\small The double pendulum setup. (a) A photograph of the double pendulum at rest, reproduced from  \citet{asseman2018learning}. (b) A simplified sketch of the setup. (c) An example of the time series extracted from the videos of the double pendulum.}
\vspace{-1em}
\end{figure}

We next validate the benefit of the expert augmentation in a controlled real-world setting. The dataset of a double pendulum introduced by \citet{asseman2018learning} contains $21$ videos of the pendulum shown in \figref{fig:pic_double_pendulum}. Each run lasts approximately $40$ seconds and is recorded at $400$Hz. We can extract the position of the pendulum limbs from each frame with elementary computer-vision tools. Each recording starts from different initial conditions, leading to many states of this chaotic system. For illustration, we showcase the evolution of the arms' angles over time in \figref{fig:double_pendulum_processed}. 

We sketch a simplified representation of the double pendulum in \figref{fig:double_pendulum_sketch}. Its state is a four-dimensional vector $y_t = \begin{bmatrix}\theta_1(t) & \theta_2(t) & \dot \theta_1(t) & \dot \theta_2(t)\end{bmatrix}^T$, containing the position and speed of both masses. We can derive (e.g., \citep{stachowiak2006numerical}) the kinetics of the frictionless pendulum from first-principle physics,\begin{align}
\ddot  \theta_1 &= \frac{-g (2 m_1 + m_2) \sin \theta_1 - m_2 g \sin(\theta_1 - 2 \theta_2) - 2 \sin(\theta_1 - \theta_2) m_2 ({\dot \theta_2}^2 l_2 + {\dot \theta_1}^2 l_1 \cos(\theta_1 - \theta_2))}{l_1 (2 m_1 + m_2 - m_2 \cos(2 \theta_1 - 2 \theta_2))},\\
\ddot \theta_2 &= \frac{2 \sin(\theta_1 - \theta_2) ({\dot \theta_1}^2 l_1 (m_1 + m_2) + g(m_1 + m_2) \cos \theta_1 + {\dot \theta_2}^2 l_2 m_2 \cos(\theta_1 - \theta_2))}{l_2 (2 m_1 + m_2 - m_2 \cos(2 \theta_1 - 2 \theta_2))}.
\end{align}
This ODE is a suitable expert model candidate for a real-world double pendulum.

We assume that $m_1 = m_2$. Therefore the effect of masses reduces to constant values in the expert ODE. The length of the two arms are known, $l_1 = 91mm$ and $l_2 = 70mm$. The total energy of the double pendulum decreases over time in all videos, which lets us speculate about frictions, not explained by the expert model. In addition, the expert model does not consider potential vibrations or errors in extracting the arms' positions. Hybrid learning has the potential to correct these mispecifications automatically. In comparison, the characterisation of the frictions from first-principle physics is challenging and is still a research subject~\citep{aghili2020energetically}.

Similarly to the damped pendulum, we consider the initial angular positions, $\theta_1(t=0)$ and $\theta_2(t=0)$, known. The encoder must predict the initial angular speeds $z_e := \{\dot \theta_1(t=0), \dot \theta_2(t=0)\}$ which are the only free parameters of the expert model.  The encoder only observes $\theta_1$ between $t=0ms$ to $t=100ms$ and $\theta_2$ between $t=50ms$ to $t=100ms$ which makes the estimation of $z_e$ complicated. Then, we predict the angular positions between $t=0$ and $t=200ms$ given $\theta_1(t=0)$ and $\theta_2(t=0)$ and the estimation of $z_e := \{\dot \theta_1(t=0), \dot \theta_2(t=0)\}$ via the hybrid decoder.
% However, in a practical setting, we might estimate the speeds at $t=5ms$ via finite difference and make backward estimation

% In contrast to the synthetic experiments in which we know the initial ODE state and the encoder predicts the parameters of the ODEs, here we only know the initial angles $\theta_1(t=0)$ and $\theta_2(t=0)$. Moreover, there are no free parameters in the ODEs. In principle, we could estimate the angular speeds via finite differences. However, the expert model would not have any parameters left, making our augmentation inapplicable. Instead, we consider the initial angular velocities as the parameters of the expert model, $z_e := \{\dot \theta_1(t=0), \dot \theta_2(t=0)$. 

We create a dataset with many initial conditions by splitting the videos into consecutive chunks of 20 frames sub-sampled at 100Hz, i.e., 200ms of video. We construct a distribution shift, as shown in \figref{fig:dis_shift_double_pendulun} from \appref{app:double_pendulum}, over the expert variables $z_e$ by splitting each $40$ seconds sequence into three parts. The training set only contains chunks from the last 16 seconds of each run. It corresponds to configurations with smaller energy and, thus, slower angular speeds than the test set, which only contains frames from the first 12 seconds. The validation set contains the remaining 12 seconds of frames in the middle.
\subsection{Results}
\begin{figure}
    \centering
    \includegraphics[width=.98\textwidth]{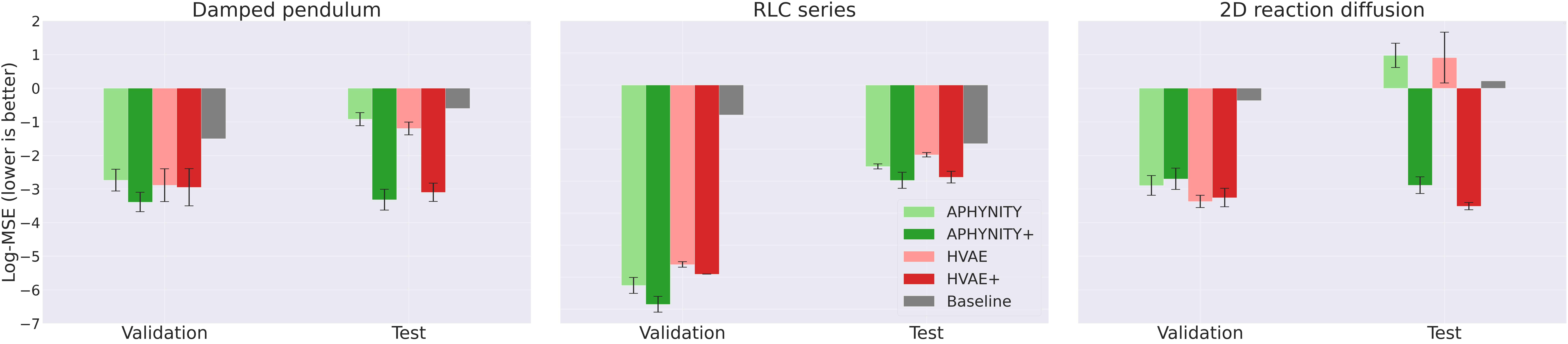}
    \caption{\small  The average log-MSEs over 10 runs for three synthetic problems on the validation and test sets. We compare HVAE (in red) and APHYNITY (in green), in light colours, to their expert augmented versions HVAE+ and APHYNITY+, in darker colours. \textit{On the test sets, AHMs outperform the original models, and by a large margin on the pendulum and diffusion problems. Moreover, augmentation conserves the relatively good performance on the validation set (IID w.r.t. the training set).}}
    \label{fig:log_mse}
\end{figure}

\paragraph{Performance gain from augmentation.}
\textit{This experiment demonstrates that HVAE and APHYNITY are not robust to OOD test scenarios in opposition to the corresponding AHMs, as shown in \figref{fig:diffusion_shift}} for the 2D diffusion problem and in \appref{app:supp_results} for the two other problems. We emphasize that our intention is not to declare a winner between HVAE and APHYNITY. Indeed, both algorithms have already demonstrated performance superior to black box ML models. Hence, we only report a very simple baseline that is the mean value of the signals. We want to compare performance in OOD settings and empirically validate the benefit of AHMs.
We compare the predictive performance in \figref{fig:log_mse} (see \tabref{tab:synth_log_mse} for the raw numbers).  Although classical hybrid learning strategies do very well on the IID validation set, they exhibit poor generalization on OOD test sets for all three problems. We also observe some disparity between APHYNITY and HVAE. In addition to different learning strategies, this is probably due to differences in the networks' architectures as they were respectively inspired from the corresponding pendulum experiment in each paper. However, even if one method may outperform the other for some problems, they both benefit from our augmentation strategy (APHYNITY+, HVAE+). Overall, the effect of augmentation goes up to dividing the test error by a factor of $e^{4.6}\approx 100$ in some cases. 

% \begin{table}[]
%     \small
%     \centering
%     \setlength{\tabcolsep}{4pt}
%     \begin{tabular}{c c|c c | c c}
%         Dataset & & APH. & HVAE & APH.+ & HVAE+ \\ \hline
% %        & Train & &  &  &  &  \\
%        \multirow{2}{*}{Pendulum} & Valid. & $6_{\pm 2}$ & $3_{\pm 1}$ & $6_{\pm 2}$ & $2_{\pm 1}$ \\
%         & Test & $66_{\pm 9}$ & $117_{\pm 10}$ & $10_{\pm 4}$ & $11_{\pm 2}$    \\ \hline
% %        & Train & &  &  &  &  \\
%        \multirow{2}{*}{RLC} & Valid. & $6_{\pm 3}$ & $38_{\pm 2}$ & $7_{\pm 5}$ & $28_{\pm 1}$ \\
%         & Test & $17_{\pm 3}$ & $25_{\pm 2}$ & $5_{\pm 2}$ & $12_{\pm 1}$ \\ \hline
        
%        \multirow{2}{*}{Diffusion} & Valid. & $2_{\pm 0}$ & $2_{\pm 0}$ & $2_{\pm 0}$ & $2_{\pm 0}$ \\
%         & Test & $27_{\pm 2}$ & $32_{\pm 10}$ & $3_{\pm 1}$ & $2_{\pm 0}$\\ \hline
%     \end{tabular}
%     \caption{Comparison of mean relative precision (in $\%$, $\pm$ indicates one standard deviation) over 10 runs of predicted physical parameters of different hybrid modelling strategies in validation and OOD test settings. Augmented versions are denoted with a $+$. \textit{While the accuracy of APHYNITY and HVAE is good on the validation set, it collapses on the OOD test set. On the opposite, the augmented versions perform well on both validation and test sets.}}
%     \label{tab:param_acc}
%     \vspace{-1.5em}
% \end{table}

\begin{figure}
\vspace{-.5em}
    \centering
    \includegraphics[width=1.\textwidth]{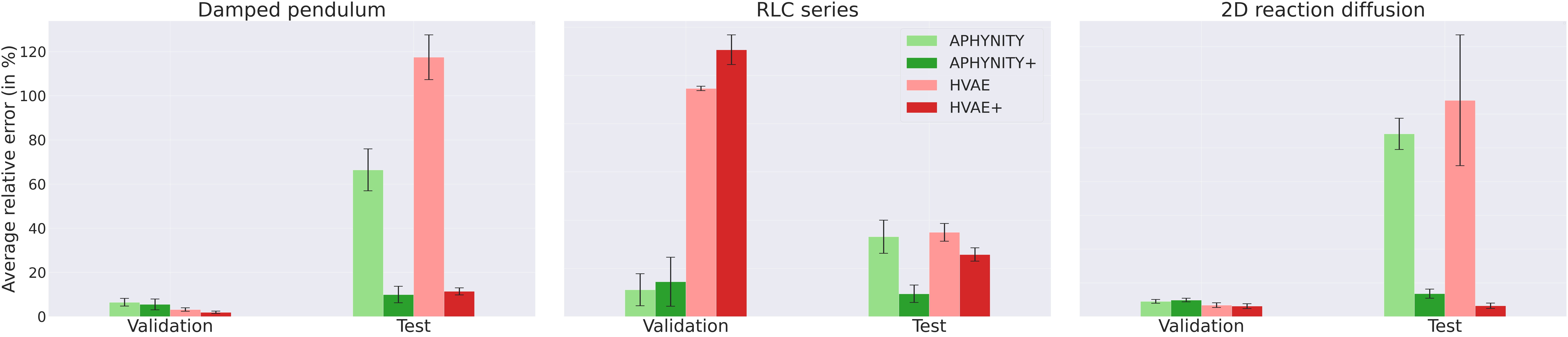}
    \caption{\small Comparison of mean relative precision (in $\%$, $\pm$ indicates one standard deviation) over 10 runs of predicted physical parameters of different hybrid modelling strategies in validation and OOD test settings. Augmented versions are denoted with a $+$. \textit{While the accuracy of APHYNITY and HVAE is good on the validation set, it collapses on the OOD test set. On the opposite, the augmented versions perform well on both validation and test sets.}}
    \label{fig:param_acc}
\end{figure}

\paragraph{Stability for non-exact models.}
The empirical results from \figref{fig:log_mse} are very important as they show that even when the decoder is not $\Omega\textit{-exact}$ (and hence not $\accentset{+}{\Omega}\textit{-exact}$), augmentation may still work. In particular, \figref{fig:param_acc} shows that the encoder does not predict the physical parameters perfectly. This indicates that the encoder is not $\Omega\textit{-exact}$ and neither should be the decoder. This plot shows the relative error on the physical parameters computed as $\sum_{i=1}^k\frac{1}{k}\left|\frac{z_e^i - \mu_\theta^i}{z_e^i}\right|$, where $\mu_\theta^i$ is the estimated most likely value of the $i^{th}$ component of the physical parameters. We first notice that APHYNITY and HVAE perform differently and their performance depends on the specific problem. While APHYNITY accurately estimates the physical parameters on the IID validation set for the 3 problems, HVAE's performance are mixed on the RLC problem as it makes prediction that are around $120\%$ away from the nominal parameter value on average whereas APHYNITY reduces this error to $6\%$. Interestingly, we observe that the proposed augmentation strategies improve the encoder such that it accurately estimates the physical parameters also on the OOD test set even for HVAE on the RLC problem. This confirms that the augmentation strategy is helpful even when the hybrid model is not $\Omega\textit{-exact}$. As a conclusion, augmented hybrid learning outperforms classical hybrid learning both on the predictive accuracy and at inferring the expert variables.

\begin{figure}[H] 
\begin{minipage}[c]{0.5\textwidth}
\centering
        \includegraphics[width=1.\textwidth]{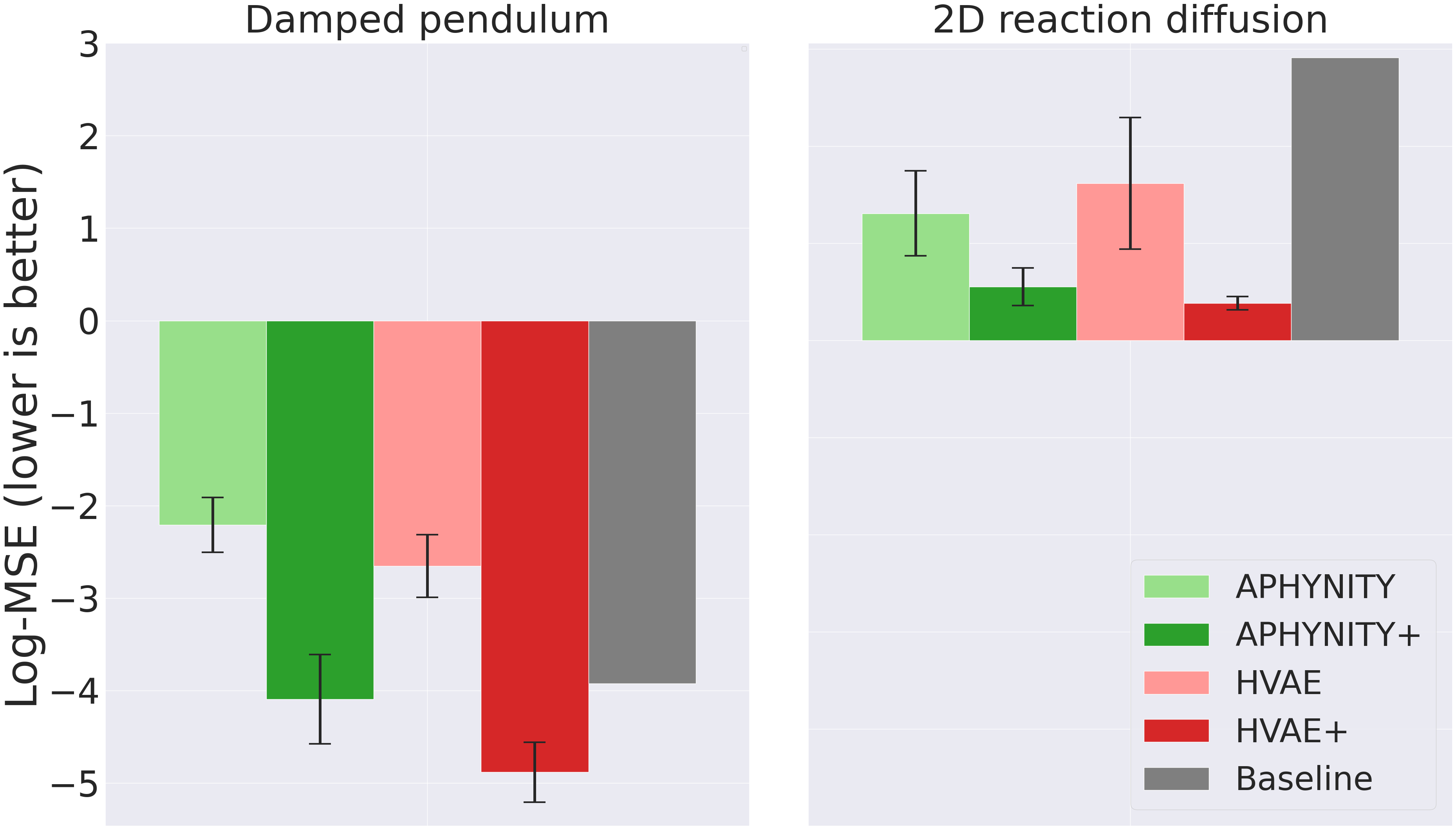}
    \caption{\small The average log-MSEs over 10 runs for the \textit{damped pendulum} and \textit{2D reaction diffusion} problems on a test distribution for which $z_a$, in addition to $z_e$, is also shifted. \textit{AHM achieves better peformance than standard algorithms even when the test distribution support $z_a$ differs from the training.}}
    \label{fig:za_shifted}
\end{minipage} \hfill
\begin{minipage}[c]{0.45\textwidth}
\centering
        \includegraphics[width=.9\textwidth]{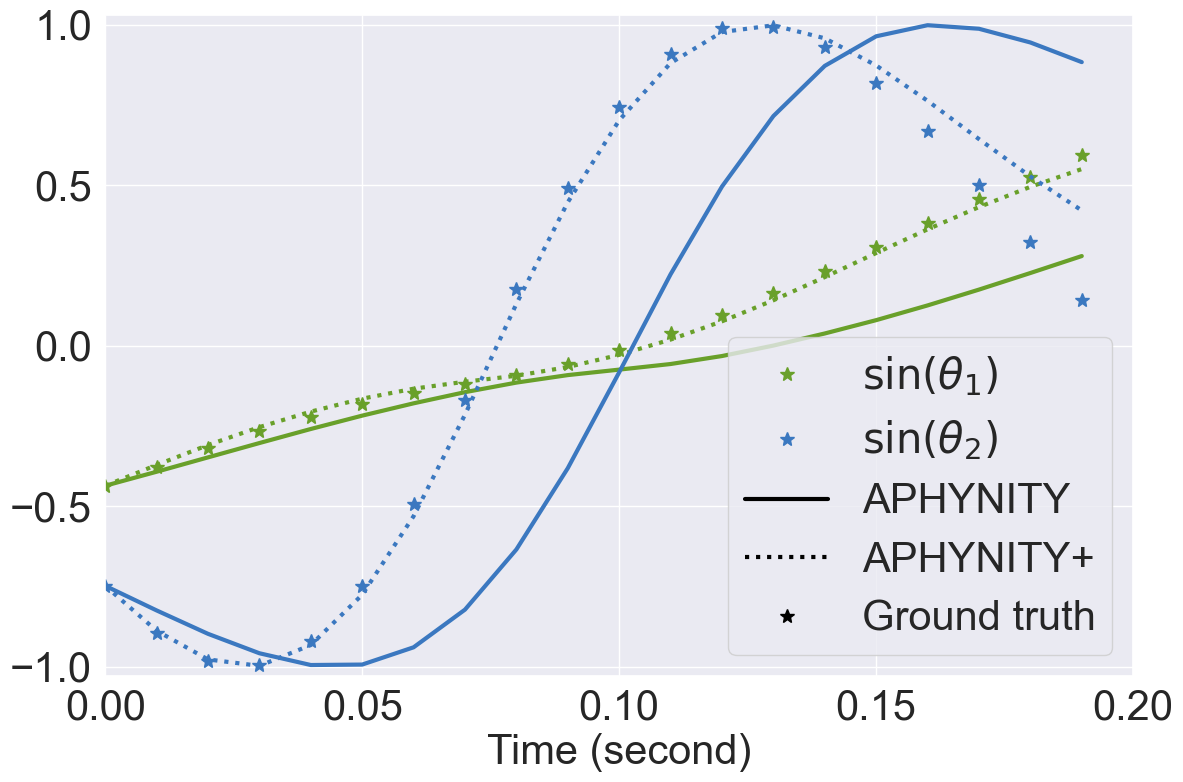}
        \vspace{-.5em}
    \caption{
    {\small A cherry-picked example of the predicted angular positions of the double pendulum. \textit{We observe that the proposed expert augmentation allows the hybrid model to predict more accurately the state of the double pendulum in the future than the non-augmented hybrid model.}}}
    \label{fig:example}
\end{minipage}
\vspace{-2em}
    \end{figure}

\paragraph{Effect of out of expertise shift.}
\textit{This experiment supports that our augmentation strategy may remain beneficial even when the train and test supports of $z_a$ are not identical.} This scenario corresponds to samples $(x, y)$ generated by $(z_a, z_e) \in (\accentset{\ast}{\mathcal{Z}}_a \setminus \mathcal{Z}_a) \times \tilde{\mathcal{Z}}_e$ depicted by the violet domains in \figref{fig:ood_vizu}. In \figref{fig:za_shifted} we observe the log-MSE of augmented and non-augmented hybrid models trained for $(z_a, z_e) \in \mathcal{Z}_a \times \mathcal{Z}_e$ on test data that are generated with $(z_a, z_e) \in \tilde{\mathcal{Z}}_a \times \tilde{\mathcal{Z}}_e$. For the pendulum, the support over $z_a=\alpha$ is $\left[0, 0.3\right]$ in train and $\left[0.3, 0.6\right]$ in test; For the 2D reaction diffusion, $z_a=k$ is $\left[0.003, 0.005\right]$ in train and $\left[0.005, 0.008\right]$ in test. We observe that augmented models outperform the original models by a large margin. These results suggest that augmentation is valuable even when the distribution shift is not caused by the expert variables. However, if the shift on $z_a$ becomes the dominant effect, augmented models also eventually becomes vulnerable to shifts on $z_e$ as demonstrated by supplementary experiments in \appref{app:exp_details}.

\begin{figure}[H]
    \centering
    \begin{subfigure}[b]{0.48\textwidth}
        \includegraphics[width=1.\textwidth]{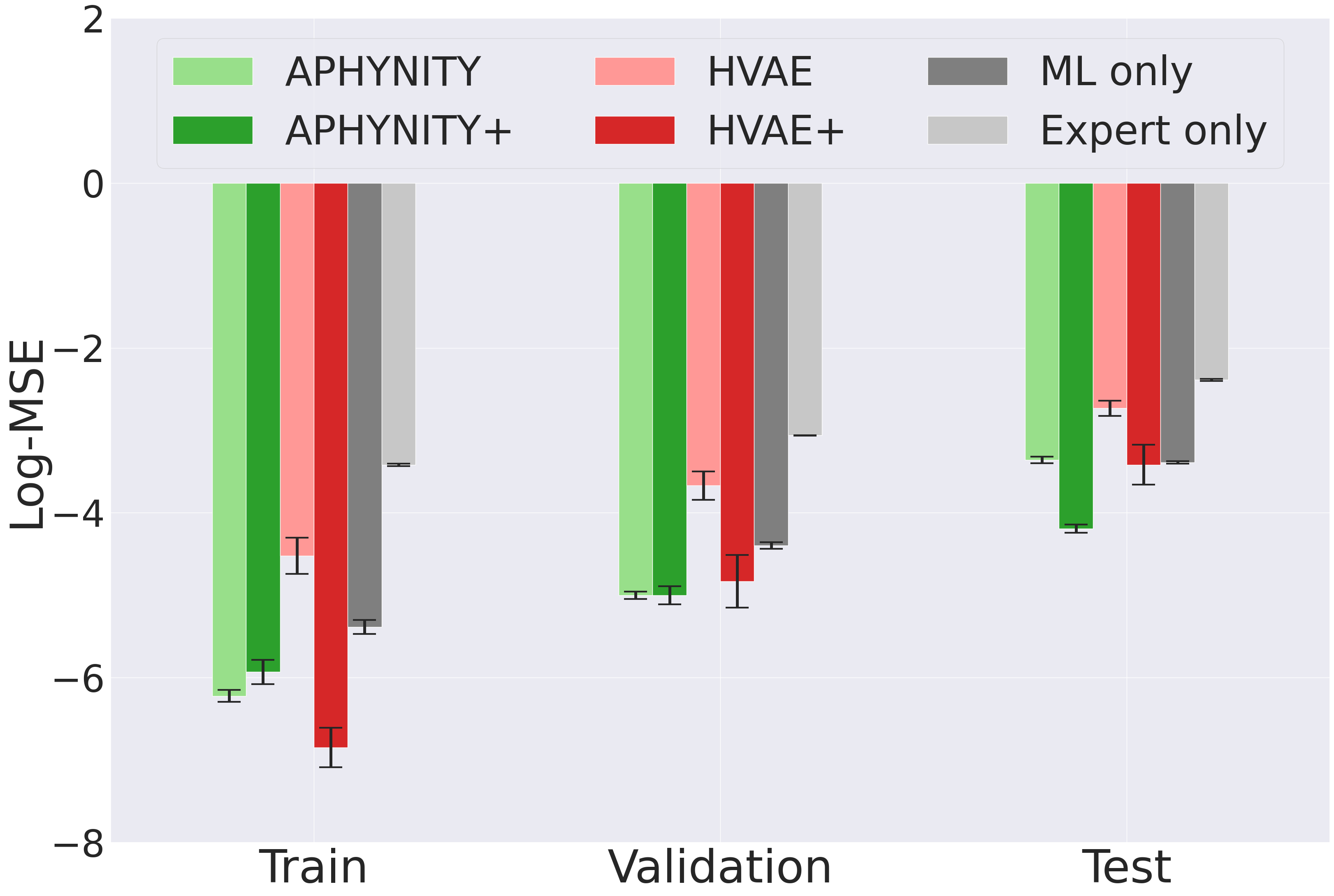}
        \caption{}\label{fig:double_pendulum_mse}
    \end{subfigure}  
    \hfill
    \begin{subfigure}[b]{0.48\textwidth}
        \includegraphics[width=1.\textwidth]{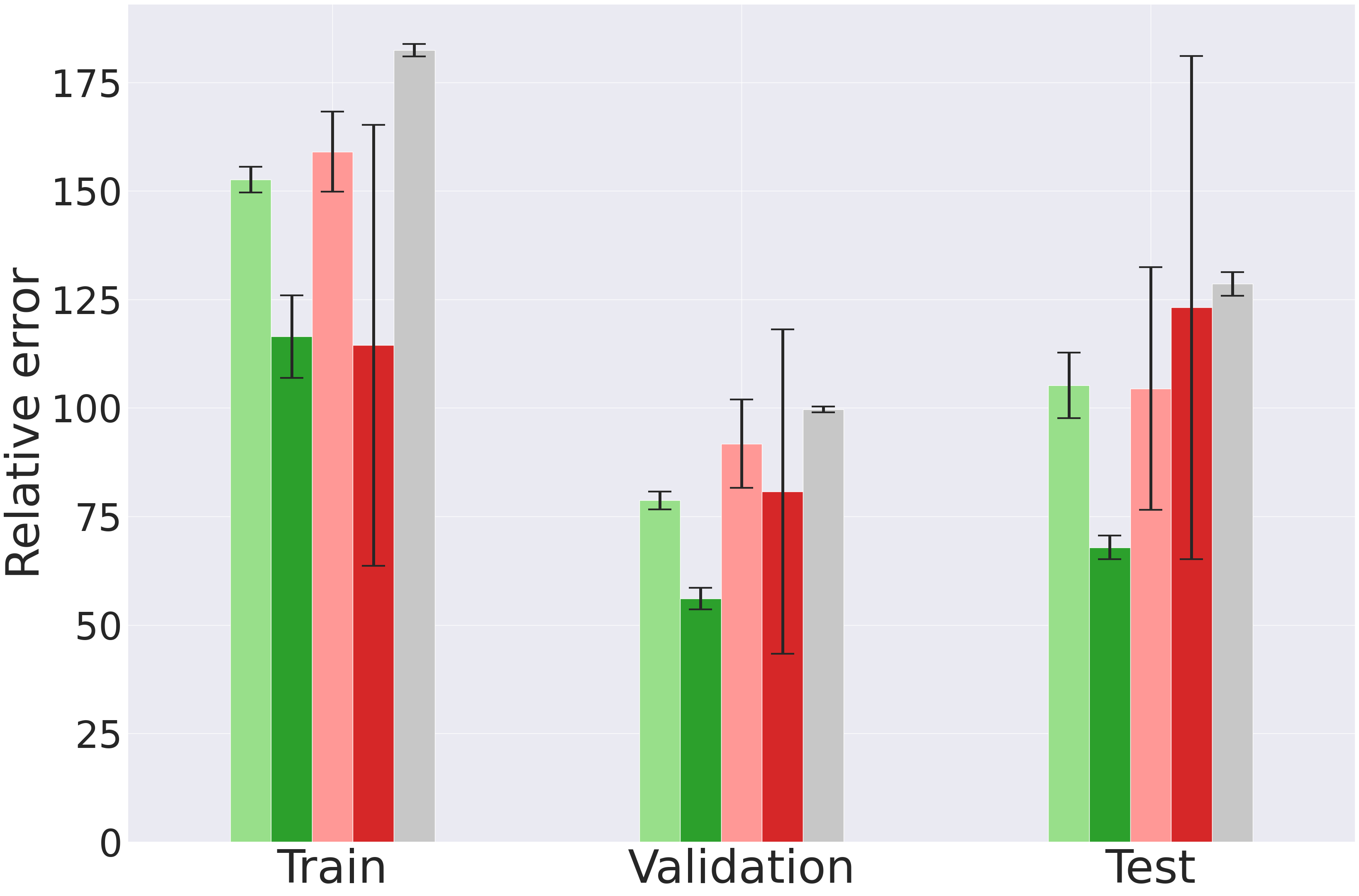}
        \caption{}\label{fig:double_pendulum_accuracy}
    \end{subfigure}  
   \caption{\small The results of the double pendulum experiment. (a) The average log-MSEs over three experiments. The baselines rely either only on the expert ODE or a neural ODE to predict the pendulum's state. \textit{The proposed expert augmentation slightly reduces the predictive performance on the training set for APHYNITY but increases the generalisation capabilities of both APHYNITY and the HVAE. APHYNITY+ outperforms the baselines on all sets.} (b) The average relative errors on the initial angular speeds over three runs. \textit{The proposed expert augmentation improves the accuracy of the physical parameters estimation both in the IID and OOD settings for APHYNITY.}} \label{fig:double_pendulum_results_bis}
\end{figure}

\paragraph{Real-world double pendulum.}
\textit{This experiment demonstrates the potential effectiveness of expert augmentation for real-world applications} 
 In \figref{fig:double_pendulum_results_bis}, we compare the empirical performance of APHYNITY(+) and HVAE(+) with two baselines: 1) \textit{Expert only}: the ODE of a friction-less double pendulum, 2) \textit{ML only}: an agnostic neural ODE.   We observe in \figref{fig:double_pendulum_mse} that the augmentation improves the validation and test predictive performance by a non-negligible margin, as confirmed visually by \figref{fig:example}. In addition, the augmentation improves the estimation of the expert parameters by up to a factor of two for APHYNITY in the OOD test scenarios, as shown in \figref{fig:double_pendulum_accuracy}. In order to achieve these results, we finetune the encoder on expert augmentations with $\dot \theta_1 \sim \mathcal{U}\left[ -15, 15\right]$ and $\dot \theta_2 \sim \mathcal{U}\left[ -30, 30\right]$. 

 We have observed that optimizing the HVAE on the double pendulum data eventually becomes numerically unstable along training. It is why we did not manage to obtain HVAE's performance on par with APHYNITY and the ML only baseline. In consequence, the expert augmentation is also unstable if it relies on the best model. We circumvent these numerical issues by applying the expert augmentation to an earlier version of the HVAE model. The expert augmented model HVAE+ outperforms the best HVAE model on predicting the trajectory as seen in \figref{fig:double_pendulum_mse}. Finally, we study the effect of applying the expert augmentation at various stages during training for APHYNITY in \appref{app:add_results_double_pendulum}. We observe that expert augmentation may reduce the gap between the train/validation/test performance on the double pendulum even when the interaction model is not fully trained. This result hints again that expert augmentation can help even when the interaction model is not learned perfectly. 

%%%%%%%%%%%%%%%%%%%%%%%%%%%%%%%%%% 5 Related work %%%%%%%%%%%%%%%%%%%%%%%%%%%%%%%%%%
\section{Related work}
\subsection{Hybrid modelling}
Hybrid Learning, or gray box modelling as called in its early days in the 90's \citep{old_hl_1, old_hl_2, old_hl_3, old_hl_4, old_hl_5}, has been a popular method to learn models that are both expressive and interpretable, while also allowing them to be learnt on fewer data. The interest for hybrid learning \citep{hl_1, hl_2, hl_3, hl_4, hl_5, hl_6, metnet2} has greatly increased since the outbreak of recent neural network architectures that simplify the combination of physical equations within ML models. As an example, Neural ODEs~\citep{NODE} and convolutional neural networks~\citep[][CNN]{lecun_cnn} are privileged architectures to work with dynamical systems described by ODEs or PDEs. While most of the literature focus on the predictive performance of hybrid models, recent work have also shown that this framework helps to infer the physical parameters accurately \citep{APHYNITY, HVAE}. This is aligned with \citet{nuisance_para} (see Section 40.2.2.2) which observe that inference on incomplete models results in a \textit{systematic bias}. Similar to hybrid learning, they extend the model with \textit{nuisance} parameters in order to improve its fidelity, and to reduce the systematic bias.

In this work, we decided to study \citet{APHYNITY} and \citet{HVAE} for two reasons that distinguish them from the rest of the literature. First, these are notable examples of algorithms that can be applied to a broad class of problems in contrast to papers that focus on specific applications \citep{hl_2,hl_3}. Second, those methods also learn a reliable inference model for the physical parameters, suggesting that the expert model is used properly in the generative model, which is a key assumption for our augmentation. While \citet{HVAE} claim to achieve robustness, we argue that this statement is incomplete as HVAE fails in OOD settings. In particular, their approach is only able to generalize with respect to unseen time or initial state if the model correctly identifies the latent variables $z_a, z_e$. HVAE cannot generalize to new physical parameters because the encoder's validity is bound to the training set for the physical parameters.

\subsection{Combining hybrid modelling and data augmentation}
Close to our idea is the one proposed in \citet{gan_hl} where they train a GAN model that improves the realism of a simulated image while conserving its semantic content (e.g., eyes colour) as modeled by the simulation parameters. The generated data with their annotations may then be used for a downstream task, such as inferring the properties of real images that corresponds to simulation parameters. The GAN objective from \citet{gan_hl} requires that the two distributions induced by the semantic content of real and simulated data are identical. On the opposite, we consider training data that corresponds to expert parameters with limited diversity, and overcome this scarcity with expert augmentation. Another line of work similar to ours is Sim2Real, which considers the task of transferring a model trained on simulated data to real world \citep{sim2real-1, sim2real-2, sim2real-3}. Robust hybrid learning, as a way to enhance simulations, could be used for Sim2Real.

%\subsection{Robust ML}
\subsection{Robust ML and invariant learning}
Various statistical methods have been introduced to ensure models generalize under distribution shift. Domain-adversarial objectives aimed at learning (conditionally) invariant predictors \citep{ganin2016domain,zhang2017aspect,li2018deep}, GroupDRO \citep{groupDRO_ICLR} optimizing for worst-case loss over multiple domains and IRM~\citep{IRM} as well as sub-group calibration \citep{wald2021calibration} aiming to satisfy calibration or sufficiency constraints to learn features invariant across domains. Extensions, able to infer domain labels from training data have been proposed as well \citep{lahoti2020fairness,creager2021environment}, partially inspired by fairness objectives \citep{hebert2018multicalibration,kim2019multiaccuracy}. In contrast to AHM, all of these methods rely on the variation of interest being present in the training data.

%%%%%%%%%%%%%%%%%%%%%%%%%%%%%%%%%% 6 Discussion %%%%%%%%%%%%%%%%%%%%%%%%%%%%%%%%%%
\section{Discussion}
We now discuss the potential limitations of our method and its underlying assumptions.
\paragraph{Erroneous interaction model.}
The exactness of the hybrid component $p_\theta(y|x, y_e, z_a)$ is a critical assumption underlying our expert-based augmentation strategy. Unfortunately, this component is learned from training data only. Hence, we cannot prove its exactness on the test domain in the general case as the input domain over $\mathcal{Y}_e$ may be different from the training. However, we argue that assumptions on the class of interaction model may alleviate this problem. As an example, we might consider to chose the best interaction model over a fixed set of potential interaction. If the correct interaction is present in this set we should eventually select it from data. A less extreme example is when we consider an additive hybrid model and embed this hypothesis into the interaction model, generalization to unseen $y_e$ follows as long as the range of values is the same as in the training set. If this assumption is too strong, we could still expect that $p_\theta(y|x, y_e, z_a)$ generalizes to unseen $y_e$ because hybrid learning drives $y$ samples from $p_\theta$ to be close to $y_e$. It implies that the corresponding function approximator is stable, which helps generalizing to unseen scenarios. However, we must acknowledge that we cannot guarantee the stability of the interaction model in the general case. In practice, finding a good inductive bias for the interaction model is important. Moreover, it is usually easier to embed an effective inductive bias in the interaction model rather than in the encoder. Indeed, the parameter identification (the encoder $q_\psi$) is often less-well understood or more complex than the generative model itself.
\paragraph{Diagnostic.}
While crucial, we cannot guarantee the exactness of the decoder $p_\theta$ in general because we only evaluate the encoder and the decoder jointly on data points $(x, y, x_o, y_o)$. However, in some cases we can detect model misspecification by observing that the predictive model $p_{\theta, \psi}(y|x, x_o, y_o)$ is imperfect. Making this observation is not always simple as it requires prior knowledge on the expected accuracy of an exact model. However, when the system is identifiable, we may argue that the accuracy should be only limited by the intrinsic measurement noise on $y$.

\paragraph{Relaxing exactness.}
Even with a solid inductive bias on the decoder, achieving exactness is hard in practical settings. However, our experiments demonstrate that expert-augmentation works in practice. We can explain this by looking at \figref{fig:ood_vizu}. If the generative model that maps $x$ and $(z_a, z_e)$ is incorrect, the mapping from $\mathcal{Z}_a$ and $\mathcal{Z}_e$ could be slightly off from $\accentset{+}{\Omega}$. However, this does not preclude the set of augmented samples from being closer to $\accentset{+}{\Omega}$ than $\Omega$ and from inducing a better predictive model on $\accentset{+}{\Omega}$ than the original model trained only on $\Omega$. Another argument is the effectiveness of data augmentation for training classical deep learning models, which works well even when some augmentations do not generate realistic samples. In addition, \citet{zhanggeneralization} have observed that the dominant cause of overfitting in classical variational autoencoders is usually coming from the encoder, similarly to what happens with amortized hybrid learning. In this regard, an unexpected advantage of expert augmentation would be to reduce this source of overfitting in addition to the OOD robustness gain.

\paragraph{Limitations.}
We have considered expert models that are parameterized by a small number of parameters and are covered densely via sampling. For higher dimensional parameter space the augmentation strategy might become inapplicable. Hence, a more ingenious sampling strategy, such as worst-case sampling, would be required. Another difficulty is choosing a plausible range of parameters that contains both the train and the test support; this will often need a human expert in the loop. If the chosen distribution does not cover some test configurations the robustness guarantees for these configurations collapse. On the opposite, a distribution that is too broad may also impact negatively the quality of the model. For instance, learning the right encoding behaviour for those unnecessary configurations consumes representation capacity that is then unavailable for IID configurations. In this case the model might perform worse than before augmentation on the training distribution. Indeed, some of our results have shown that the training error may slightly increase while the test error decreases.

In addition, we assume that the train distribution of $z_a$ is representative of the test distribution. We empirically observed that a softer version of this assumption could be enough. However, performance will eventually decline as the support of the test distribution for $z_a$ is far from the training domain. Finally, we have only validated our expert augmentation for amortized inference settings. Nevertheless, online inference algorithms, such as Markov-chain Monte Carlo, also rely on hyperparameters~\citep{campbell2021gradient} or learnt distributions~\citep{brofos2022adaptation} which must be tuned to the specific problem of interest. They could also benefit from expert augmentation if those actionable components are contextually predicted by a ML model.
%%%%%%%%%%%%%%%%%%%%%%%%%%%%%%%%%% 7 Conclusion %%%%%%%%%%%%%%%%%%%%%%%%%%%%%%%%%%
\section{Conclusion}
We have described hybrid learning within a probabilistic model in which one component of the latent process, denoted the expert model, is known. In this context, we have established that state-of-the-art algorithms are vulnerable to distribution shifts. Grounded in this formalisation, we have derived that expert augmentations induce robustness to OOD settings. We have shown on a set of synthetic settings and on a real-world system the favorable effect of expert augmentation on the OOD performance where standard hybrid learning algorithm fails. Finally, we have discussed how our assumptions transfer to real-world settings and have described potential shortcomings. 
% An experiment with a real-world system showed that the proposed strategy may improve existing hybrid learning algorithms in non-artificial settings even though some assumptions upon which our augmentation strategy builds may be violated in practice. 
%Our experiments evidence that expert augmentations may be beneficial even when assumptions on the class of distribution shift do not hold.

Our augmentation should benefit from future progress in hybrid learning as it shall apply to any amortized hybrid modelling algorithms. Providing more substantial constraints on the targeted hybrid model is an essential direction for further improving the robustness of hybrid models. For instance, the minimal description length principle \citep{mdl_book} could be an excellent resource for investigating the balance between the model's capacity and robustness. In parallel, empirical and theoretical investigations of the learnt interaction model's properties in various practical settings would unlock new understandings of the capabilities and limitations of hybrid learning algorithms. In the context of expert augmentation, it should enable to do more realistic assumptions for its applicability and to derive more precisely the expected robustness gain. Future work is also needed to show how hybrid learning and expert augmentation translate into performance gain over classical ML on challenging real-world applications.

%%%%%%%%%%%%%%%%%%%%%%%%%%%%%%%%%% Acknowledgements %%%%%%%%%%%%%%%%%%%%%%%%%%%%%%%%%%
% \section*{Acknowledgements}
% We would like to acknowledge Andy Miller, Dan Busbridge, Jason Ramapuram, Joe Futoma, and Mark Goldstein for providing useful feedback on this manuscript or an earlier version of it. 

\bibliography{main}
\bibliographystyle{icml2022}

\clearpage
\onecolumn
\appendix
\section{Additional description of expert augmentation}\label{app:ahm_desc}
We provide the procedure to do expert augmentation for robust hybrid learning as the sequence of steps below.
\begin{enumerate}
    \item Train both the encoder $q_\psi(z_a, z_e|x, y)$ and the interaction model $p_\theta(y|x_o, z_a, y_e)$ with a hybrid learning algorithm, by minimizing the corresponding loss $\mathcal{L}(\psi, \theta) = \mathbb{E}_{\mathcal{D}}\left[\ell(x, y; \theta, \psi)\right]$ on the training set $\mathcal{D}$ (see line $8$ in Algorithm~\ref{alg:exp_augm});
    \item Decide on an augmented distribution $\accentset{+}{p(z_e)}$ for $z_e$ that contains both train and test scenarios (see line $17$ in Algorithm~\ref{alg:exp_augm});
    \item Reproduce the following steps to generate a dataset $\accentset{+}{\mathcal{D}}$ of observations and expert variables $(x, y, z_e) \sim \mathbb{E}_{p(z_a)p(y_e|z_e, x, y)}\left[ p(z_e) p(x) p_\theta(y|y_e, z_a, x)\right]$ (see lines $15$ to $21$ in Algorithm~\ref{alg:exp_augm}):
    \begin{enumerate}
        \item Sample $(x_o, y_o)$ from the data;
        \item Sample $z_a$ from the posterior $q_\psi(z_a|x_o, y_o)$;
        \item Sample $z_e$ from $p_{+}(z_e)$;
        \item Push forward $x, z_a$ and $z_e$ in the generative model as $y_e \sim p(y_e|x_o, z_e)$ and $y \sim p_\theta(y|x_o, z_a, y_e)$;
        \item Add the triplet $(x_o, y, z_e)$ to the augmented training set $\accentset{+}{\mathcal{D}}$.
    \end{enumerate}
    \item Freeze the interaction model, and fine-tune the encoder $q_\psi(z_a, z_e|x, y)$ on the augmented dataset $\accentset{+}{\mathcal{D}}$ by minimizing $\accentset{+}{\mathcal{L}}(\psi, \theta) = \mathbb{E}_{\accentset{+}{\mathcal{D}}}\left[\ell(x, y; \theta, \psi) - \log q_\psi(z_e|x, y)\right]$ (see line $10$ in Algorithm~\ref{alg:exp_augm}).
\end{enumerate}
\begin{algorithm}[H]
\begin{algorithmic}[1]
	\caption{Expert augmented hybrid learning} \label{alg:exp_augm}
        \State $\mathcal{D}:= \{(x^{(i)}, y^{(i)})\}_{i=1}^{N} \in (\mathcal{X} \times \mathcal{Y})^{N}$  \Comment{A training set}
        \State $q_\psi(z_a, z_e | x, y)$ \Comment{A parametric encoder}
        \State $p (y_e| x, z_e)$ \Comment{An expert model}
        \State $p_\theta (y| x, y_e, z_a)$ \Comment{A parametric decoder}
        \State $l(x, y, \theta, \psi)$ \Comment{A hybrid learning objective function}
        \State $p_{+}(z_e)$ \Comment{A prior distribution on $z_e$ that covers both train and test scenarios}
    \Procedure{Training}{}      
    \State $\psi^\star, \theta^\star \gets \arg\min_{\psi, \theta} \mathbb{E}_{(x, y) \sim \mathcal{D}} \left[l(x, y, \theta, \psi) \right]$
    \State $\accentset{+}{\mathcal{D}} \gets \Call{GenerateAugmentedSet}{}$
    \State $\psi^\star \gets \arg\min_{\psi} \mathbb{E}_{(x, y, z_e) \sim \accentset{+}{\mathcal{D}}} \left[l(x, y, \theta^\star, \psi) - \log q_\psi(z_e|x, y)\right]$
    \State \Return $\psi^\star, \theta^\star$
    \EndProcedure
    \Procedure{GenerateAugmentedSet}{} 
    \State $\accentset{+}{\mathcal{D}} \gets \{\}$
    \ForEach{$(x_o, y_o) \in \mathcal{D}$}
        \State $z_a \sim q_{\psi^\star}(z_a, z_e | x_o, y_o) $
        \State $z_e \sim p_{+}(z_e)$
        \State $y_e \sim p(y_e| x, z_e)$
        \State $y \sim p_{\theta^\star} (y| x, y_e, z_a)$
        \State $\accentset{+}{\mathcal{D}} \gets \accentset{+}{\mathcal{D}} \cup \{(x_o, y, z_e)\}$
    \EndFor
    \State \Return $\accentset{+}{\mathcal{D}}$
        \EndProcedure

\end{algorithmic}

\end{algorithm}

\section{Additional details on the hybrid-VAE} \label{app:HVAE}
We now provide the definition of the different regularizers employed by the hybrid-VAE. Our definition slightly differs from \citet{HVAE} as we explicitly accounts for known input variables $x$ and thus consider a dataset of pairs $(x, y)$. Although our framework should hold for non-deterministic expert models, we have directly employed the regularizers proposed by \citet{HVAE} which assume a deterministic expert models. We encourage the non-familiar reader to check the original work on hybrid-VAE for a thorough discussion of the effect of the different regularizers.

The first regularizer $R_{PPC}$ aims to inhibit unnecessary flexibility of the learned parts of the decoder. To this purpose, \citet{HVAE} proposes to minimize the Kullback-Leibler divergence between the posterior predictive distributions of the hybrid model and the expert model:
$$\mathbb{KL}\left[ p_{\theta, \psi}(y | \mathcal{D}) || p^e_{\theta, \psi}(y | \mathcal{D}) \right],$$
where
$$
p_{\theta, \psi}(y | \mathcal{D}) = \int p_{\theta}(y | x_o, z_a, z_e) q_{\psi}(z_a, z_e | x_o, y_o) p_{\mathcal{D}}(x_o, y_o) dx_o dy_o dz_a dz_e,
$$
and
$$
p^e_{\theta, \psi}(y | \mathcal{D}) = \int p_{\theta}(y_e | x_o, z_e) q_{\psi}(z_e | x_o, y_o) p_{\mathcal{D}}(x_o, y_o) dx_o dy_o dz_e .
$$
As the posterior predictive distributions are intractable, \citet{HVAE} introduce a lower bound on it as
$$R_{PPC} = \mathbb{E}_{x, x_o, y_o \sim p_\mathcal{D}}\left[ \mathbb{E}_{q(z_a, z_e|x_o, y_o)}\mathbb{KL}\left[ p_\theta(y | x, z_a, z_e) || p_\theta(y_e | x, z_e) \right] + \mathbb{KL}\left[ q_{\psi}(z_a, z_e | x_o, y_o) || p(z_a) p(z_e) \right] 
\right].$$

The second regularizer $R_{DA, 1}$ helps the two-steps encoder of the hybrid-vae to be grounded into the physics. The first encoder should map the input to signals that can be produced by the expert model. Formally, \citet{HVAE} propose to enforce this with 
$$R_{DA, 1} = \mathbb{E}_{p_{\mathcal{D}}(x, x_o, y_o) q(z_a | x_o, y_o) }\left[ || g_{P, 1} (x_o, y_o, z_a) - \text{sg}(y_e(x_o, z_e=g_{P, 2}(g_{P, 1} (x_o, y_o, z_a))  ||_2^2 \right],$$
where the encoder first output a posterior $q(z_a | x_o, y_o)$ for $z_a$, then transform $y_o$ into something that should match the expert model with $g_{P, 1}$, and finally $g_{P, 2}$ output the posterior distribution over the physical parameters. The symbol sg denote the stop gradient operator.

Finally, the third regularizer helps the second second step of encoding $g_{P, 2}$ to properly encode any physical configuration. In particular, it should be able to encode properly any pair $(x, y_e)$ that we could generate with the expert model. The regularizer takes the form
$$R_{DA, 2} = \mathbb{E}_{x, z_e}|| g_{P, 2}(y_e(x, z_e)) - z_e ||_2^2.$$
\section{Additional details on experiments}\label{app:exp_details}
\subsection{Hyperparameter Search}
For the three synthetic experiments with APHYNITY, we have performed a bayesian optimization search with $4 \times 100$ trials to find this set of hyperparameters. The ranges for the search were: $\tau_2 \in \left[ 0, 10\right]$, $\lambda_0 \in \left[ 0, 20\right]$, $\text{weight decay} \in \left[ 0, 0.1\right]$, $N_{iter} \in \{ 0, \dots, 10 \}$, $\text{lr} \in \left[ 0.0005, 0.1\right]$. 

For the three synthetic experiments with HVAE, we have performed a bayesian optimization search with $20 \times 25$ trials to find this set of hyperparameters. The ranges for the search were: $\alpha \in \left[ 0, 100\right]$, $\beta \in \left[ 0, 100\right]$, $\gamma \in \left[ 0, 100\right]$, $\text{lr} \in \{0.001, 0.0005, 0.0001\}$.

From the result of these searches we have chosen the hyperparameters values provided below. Reproducing our experiments by re-generating new synthetic datasets can change the numbers. However we have observed a constant benefit of applying the expert augmentation even when the non-augmented model was not tuned perfectly.

For the double pendulum experiment we did not run any hyperparameter search and there might exist better set of hyperparameters. However, our objective was only to show that expert augmentation helps to gain performance in OOD settings, even when the experiments is not well-controlled. We believe our conclusion should not be sensitive to applying augmentation on a better fine-tuned model. Our theory suggest the opposite: if the model is better, it should be able to generate good augmented samples and gain generalization capabilities.
\subsection{Damped pendulum}
\paragraph{Datasets.}
We use Neural Ordinary Differential Equations (NODE) \citep{NODE} to solve the ODE ruling the damped pendulum. Each sample is simulated for $t_0=0s$, $t_1=5s$, and $t_2=20s$, with a time resolution equal to $0.1$ second. The models are trained with only the realizations between $t_0$ and $t_1$. At test and validation time, the pair $(x_o, y_o) = (y_0, [y_{i\Delta t}]_{i=1}^{t_1 \!/\! \Delta t})$, $x=y_{t_1}$ and the model predicts $y=[y_{i\Delta t}]_{i=t_2 \!/\! \Delta t + 1}^{t_2 \!/\! \Delta t}$. The initial angular speed is always $0$ and $\theta_0 \sim \mathcal{U}(-\frac{\pi}{2}, \frac{\pi}{2})$.

The training set is made of $1000$ samples and the validation set of 100 samples. They are both generated by sampling uniformly $z_a := \alpha$ from $\mathcal{Z}_a := \left[ 0, 0.6 \right]$ and $z_e := \omega_0$ from $\mathcal{Z}_e := \left[ 1.5, 3.1 \right]$. The shifted test set contains 100 samples generated by sampling uniformly $z_a$ in $\mathcal{Z}_a$ and $z_e$ in $\tilde{\mathcal{Z}}_e := \left[0.5, 1.5 \right]$.
\paragraph{APHYNITY.}
Our model is composed of a $1$-layer RNN with $128$ units that encodes the input signal $y_{0:t_1}$ as $h(y_{0:t_1}) \in \mathbb{R}^{128}$. An MLP with $3$ layers of $150$ units and ReLU activations maps $h$ to $\mathbb{R}_+$ to predict $\omega_0$. The function $f_a: \mathbb{R}^{128} \times \mathbb{R}^2$ is an MLP with $3$ layers of $50$ units and ReLU activations (no activation for the last layer). The models are trained for $50$ epochs with Adam with no weight decay and a learning rate equal to $0.0005$. For the Lagrangian optimization we use $N_{iter}=5, \lambda_0=10, \tau_2 = 5$ (see \citep{APHYNITY}. The augmented data are generated by sampling uniformly $z_e \in \accentset{+}{\mathcal{Z}}_e := \left[0.5, 3.5\right]$ and $z_a$ from the marginal predictive prediction of the model, that is we use the training dataset to infer values of $z_a$ and use these as samples. The batch size is $100$.

\paragraph{HVAE.}
We use the notations from \citet{HVAE} to describe the architecture of the VAE. The network $g_{p, 1}: \mathbb{R}^2 \times \mathbb{R}^{d_a}$, where $d_a=1$ is the size of the latent space for the interaction model, is supposed to filter the observations so that they can be generated by the expert model. It has $2$ hidden layers with $128$ units, $g_{p, 2}$ is an MLP with the following hidden layers $\left[128, 128, 256, 64, 32\right]$ and takes the full sequence of filtered states and predicts the mean and variance of a normal distribution that parameterize the posterior $p_\theta(z_e|x, y, z_a)$. Another network, $g_a$ takes the sequence of observations and predict the posterior distribution of $z_a$ as a normal distribution. This network has the following hidden layers $\left[256, 256, 128, 32\right]$. All networks have SeLU activations. In general the decoder of HVAE can be anything that combines the expert model in order to produce samples in the observation space, as we made the hypothesis that the ODE is just missing an additive term, the decoder is a NODE where the function is the sum of $f_e$ and $f_a$ a two hidden layers MLP with $64$ units and SeLU activation (except for the last layer that has no activation). The likelihood model is also Gaussian with the mean being predicted by the NODE and the variance learned but shared for all observations. For additional details on our architecture and implementation details we encourage the interested reader to check our code (see \url{https://github.com/apple/ml-robust-expert-augmentations}). 

The networks are trained jointly for $1000$ epochs with Adam optimizer, with a learning rate equal to $0.0005$, weight decay equal to $0.000001$ and batch size $200$. The other parameters are set to $\gamma = 1$, $\alpha=0.01$ and $\beta=0.01$. The HVAE also relies on some augmentation during training and in order to compare fairly our model to theirs we use the same distribution for our augmentation and theirs that is $z_a \sim \mathcal{N}(0, I)$ and $z_e \sim \mathcal{U}(0.5, 3.5)$. 

\subsection{RLC series}
\begin{figure}
    \centering
    \includegraphics[width=.4\textwidth]{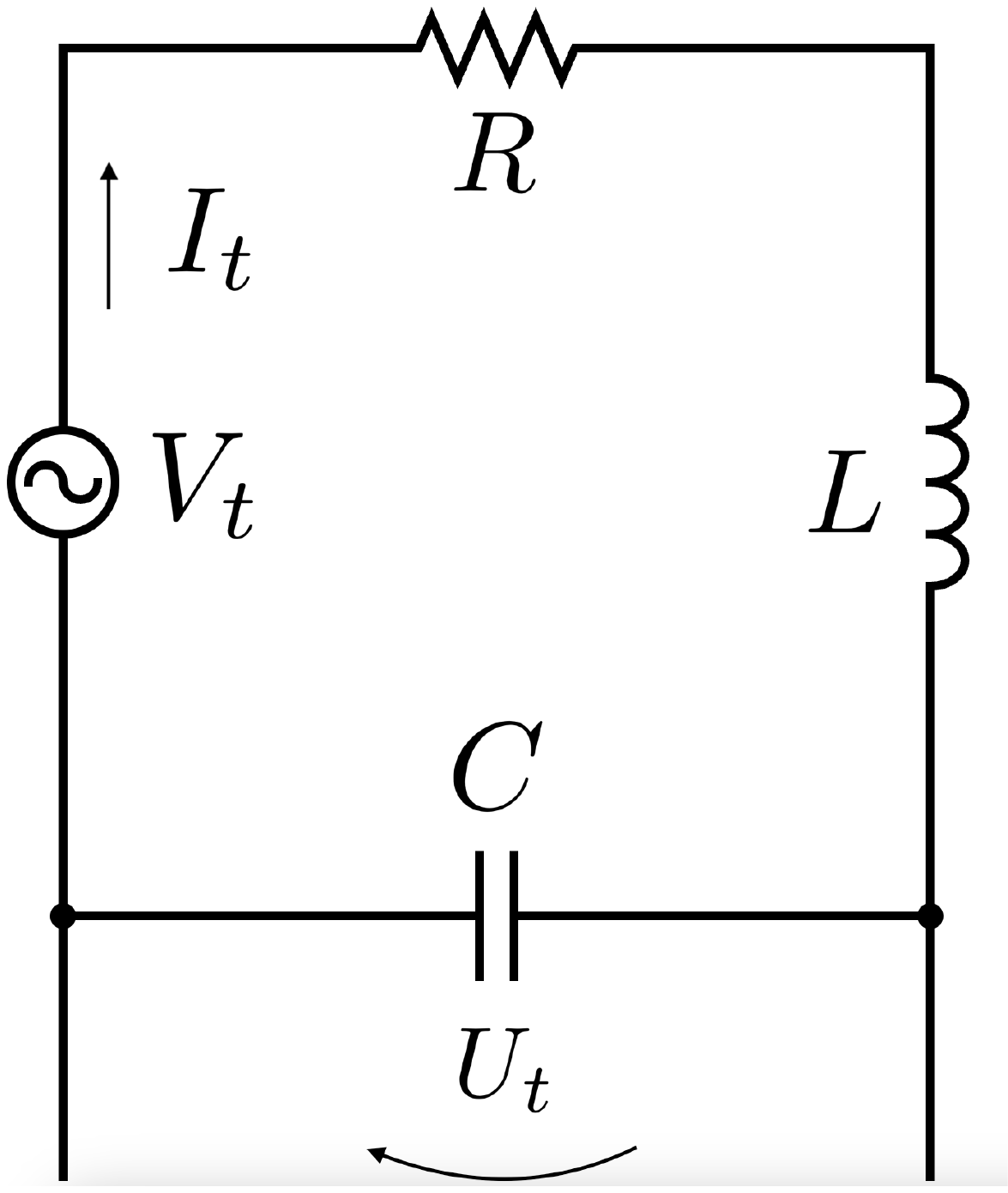}
    \caption{A generic schematic of a RLC series circuit.}
    \label{fig:RLC}
\end{figure}
\paragraph{Datasets.}
Similar to the damped pendulum, we use NODE to solve the ODE ruling the RLC circuit. Each sample is simulated for $t_0=0s$, $t_1=5s$, and $t_2=20s$, with a time resolution equal to $0.1$ second. The models are trained with only the realizations between $t_0$ and $t_1$. At test and validation time, the pair $(x_o, y_o) = (y_0, [y_{i\Delta t}]_{i=1}^{t_1 \!/\! \Delta t})$, $x=y_{t_1}$ and the model predicts $y=[y_{i\Delta t}]_{i=t_2 \!/\! \Delta t + 1}^{t_2 \!/\! \Delta t}$. In all experiments, the initial value for $U_0 \sim \mathcal{N}(0, 1)$ and $I_0=0$, the voltage source delivers a AC + DC tension $V(t) = 2.5 \sin(4 \pi t) + 1$.

The training set is made of $2000$ samples and the validation set of 100 samples. They are both generated by sampling uniformly $z_a := R$ from $\mathcal{Z}_a := \left[ 1, 3 \right]$ and $z_e := \left[L, C\right]$ from $\mathcal{Z}_e := \left[ 1, 3 \right] \times \left[ 0.5, 1.5 \right]$. The shifted test set contains 100 samples and is generated by sampling uniformly $z_a$ in $\mathcal{Z}_a$ and $z_e$ in $\tilde{\mathcal{Z}}_e := \left[ 3, 5 \right] \times \left[ 1., 2.5 \right]$.
\paragraph{APHYNITY.}
Our model is composed of a 1-layer RNN with 128 units that encodes the input signal $y_{0:t_1}$ as $h(y_{0:t_1}) \in \mathbb{R}^{128}$. An MLP with $3$ layers of $200$ units and ReLU activations maps $h$ to $\mathbb{R}_+^2$ that predicts $L$ and $C$. The function $f_a: \mathbb{R}^{128} \times \mathbb{R}^2$ is an MLP with $3$ layers of $150$ units and ReLU activations (no activation for the last layer). The models are trained for $50$ epochs with Adam with no weight decay and a learning rate equal to $0.0005$. For the Lagrangian optimization we use $N_{iter}=5, \lambda_0=10, \tau_2 = 5$ (see \citep{APHYNITY}). The augmented data are generated by sampling uniformly $z_e \in \accentset{+}{\mathcal{Z}}_e := \left[1, 5\right] \times \left[0.5, 2.5\right]$ and $z_a$ from the marginal predictive prediction of the model, that is we use the training dataset to infer values of $z_a$ and use these as samples. The batch size is $100$.

\paragraph{HVAE.}
We use the same networks' architectures than for the damped pendulum experiment. Except that $g_{p, 1}$ is has $3$ hidden layers with $100$ units.

The networks are trained jointly for $1000$ epochs with Adam optimizer, with a learning rate equal to $0.0005$, weight decay equal to $0.000001$ and batch size $100$. The other parameters are set to $\gamma = 1$, $\alpha=0.01$ and $\beta=0.01$. The HVAE also relies on some augmentation during training and in order to compare fairly our model to theirs we use the same distribution for our augmentation and theirs that is $z_a \sim \mathcal{N}(0, I)$ and $z_e \sim \mathcal{U}(1, 5) \times \mathcal{U}(0.5, 2.5)$. 

\subsection{2D reaction diffusion}
\paragraph{Datasets.}
Similar to the damped pendulum, we use NODE to solve the PDE ruling the reaction diffusion. We closely follow the experimental setting described in \citet{APHYNITY} and approximate the Laplace operator with a $3 \times 3$ discrete version of the operator. Each sample is simulated for $t_0=0s$, $t_1=1s$, and $t_2=5s$, with a time resolution equal to $0.1$ second. The models are trained with only the realizations between $t_0$ and $t_1$. At test and validation time, the pair $(x_o, y_o) = (y_0, [y_{i\Delta t}]_{i=1}^{t_1 \!/\! \Delta t})$, $x=y_{t_1}$ and the model predicts $y=[y_{i\Delta t}]_{i=t_2 \!/\! \Delta t + 1}^{t_2 \!/\! \Delta t}$. The initial state is sampled from a uniform distribution in $\left[0, 1\right]$.

The training set is made of 2000 samples and the validation set of 100 samples. They are both generated by sampling uniformly $z_a := k$ from $\mathcal{Z}_a := \left[ 0.003, 0.005 \right]$ and $z_e := \left[a, b\right]$ from $\mathcal{Z}_e := \left[ 0.001, 0.002 \right] \times \left[ 0.003, 0.007 \right]$. The shifted test set contains 100 samples and is generated by sampling uniformly $z_a$ in $\mathcal{Z}_a$ and $z_e$ in $\tilde{\mathcal{Z}}_e := \left[ 0.002, 0.004 \right] \times \left[ 0.001, 0.1 \right]$.
\paragraph{APHYNITY.}
Our model is composed of a deep CNN that encodes the input sequence of 10 images. The exact architecture can be found in the code. The dimension of $z_a$ is equal to 10. Similarly to \citet{APHYNITY} the function $f_a$ is a 3-layers CNN with ReLU activations. The models are trained for $500$ epochs with Adam with no weight decay and a learning rate equal to $0.0005$. For the Lagrangian optimization we use $N_{iter}=1, \lambda_0=10, \tau_2 = 5.$. The augmented data are generated by sampling uniformly $z_e \in \accentset{+}{\mathcal{Z}}_e := \left[0.001, 0.004\right] \times \left[0.001, 0.01\right]$ and $z_a$ from the marginal predictive prediction of the model, that is we use the training dataset to infer values of $z_a$ and use these as samples. The batch size is $100$.

\subsubsection{HVAE}
We use the notations from \citet{HVAE} to describe the architecture of the VAE. The network $g_{p, 1}: \mathbb{R}^{2\times 32 \times 32} \times \mathbb{R}^{d_a}$ is a conditional U-net, where $d_a=10$ is the size of the latent space for the interaction model, is supposed to filter the observation so that they can be generated by the expert model. The networks $g_{p, 1}$ and $g_a$ share a common backbone CNN and are, in addition, respectively parameterized by $2$ $3$-layers MLPs. All networks have ReLU activations. In general the decoder of HVAE can be anything that combines the expert model in order to produce samples in the observation space, as we made the hypothesis that the ODE is just missing an additive term, the decoder is a NODE where the function is the sum of $f_e$ and $f_a$ a $3$-layers CNN. The likelihood model is also Gaussian with the mean being predicted by the NODE and the variance learned but shared for all observations. For additional details on our architecture and implementation details we encourage the interested reader to check our code. 

The networks are trained jointly for $1000$ epochs with Adam optimizer, with a learning rate equal to $0.0005$, weight decay equal to $0.00001$ and batch size $100$. The other parameters are set to $\gamma = 1$, $\alpha=0.01$ and $\beta=0.01$. The HVAE also relies on some augmentation during training and in order to compare fairly our model to theirs we use the same distribution for our augmentation and theirs that is $z_a \sim \mathcal{N}(0, I)$ and $z_e \sim \mathcal{U}(0.001, 0.004) \times \mathcal{U}(0.001, 0.01)$. 
\subsection{Double pendulum} \label{app:double_pendulum}
\begin{figure}[h!]
    \centering
    \includegraphics[width=1.\textwidth]{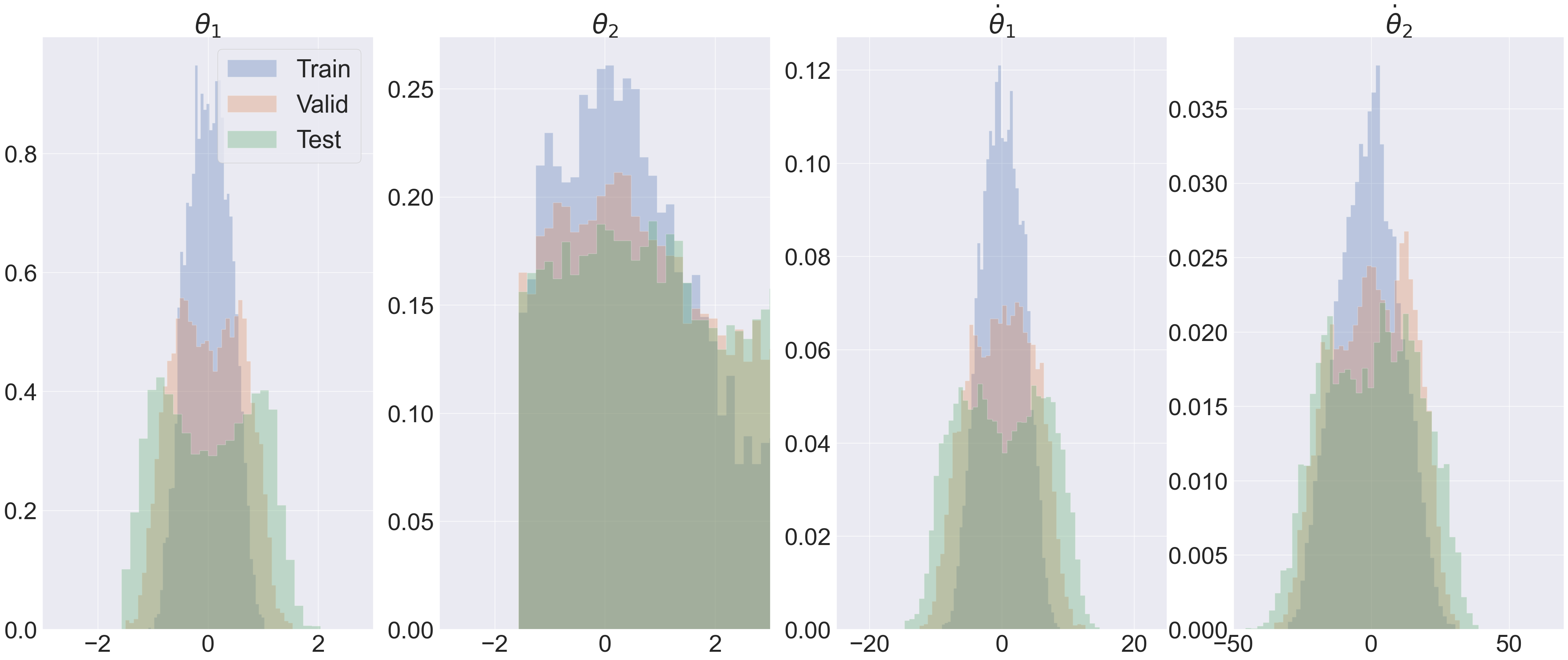}
    \caption{The distributions of initial angular positions and speeds in the experiment of the double pendulum.}
    \label{fig:dis_shift_double_pendulun}
\end{figure}

\figref{fig:dis_shift_double_pendulun} shows the marginal distributions of the angular positions and speeds in the artificially split training, validation and testing sets. We use this plot to inform the bounds of the uniform distributions we use to do the expert augmentation.

In this experiment, we only used APHYNITY because our main goal was just to show the effectiveness of the augmented version and not to compare HVAE to APHYNITY. In order to make APHYNITY works, we had to extensively search for good hyperparameters. We eventually reach the following setup.
The encoder is a fully connected neural network with $3$ layers of $300$ neurons. $F_a$ is also a fully connected neural network with $3$ layers of $300$ neurons and $z_a \in \mathbb{R}^{10}$.
The models are trained for $100$ epochs with Adam with no weight decay and a learning rate equal to $0.0005$. For the Lagrangian optimization we use $N_{iter}=1, \lambda_0=1000, \tau_2 = 5$. The batch size is $100$.

The augmented data are generated by sampling uniformly $z_e \in \accentset{+}{\mathcal{Z}}_e := \left[-15, 15\right] \times \left[-30, 30\right]$ and $z_a$ from the marginal predictive prediction of the model, that is we use the training dataset to infer values of $z_a$ and use these as samples. These bounds were chosen by looking at the distribution in \figref{fig:dis_shift_double_pendulun} .

\section{Supplementary results}\label{app:supp_results}
We now provide additional results for AHM versus standard models.

\subsection{Log-mses on the three synthetic problems}
\tabref{tab:synth_log_mse} presents the log-mses numbers on the three synthetic problems.
\begin{table}[h!]
    \small
    \centering
    \setlength{\tabcolsep}{4pt}
    \begin{tabular}{l c| c c | c c}
        Dataset & & APH. & HVAE & APH.+ & HVAE+ \\ \hline
       \multirow{2}{*}{Pendulum} & Val. & $-2.7_{\pm 0.3}$ & $-2.9_{\pm 0.5}$ & $-3.4_{\pm 0.3}$ & $-2.9_{\pm 0.6}$  \\
                            & Test   & $-0.9_{\pm 0.2}$ & $-1.2_{\pm 0.2}$ & $-3.3_{\pm 0.3}$ & $-3.1_{\pm 0.3}$  \\ \hline
       \multirow{2}{*}{RLC} & Val. & $-6.3_{\pm 0.2}$ & $-4.3_{\pm 0.1}$ & $-6.8_{\pm 0.2}$ & $-3.8_{\pm 1.5}$ \\
                            & Test  & $-2.5_{\pm 0.1}$ & $-2.2_{\pm 0.1}$ & $-3.0_{\pm 0.3}$ & $-2.1_{\pm 0.3}$\\ \hline
       \multirow{2}{*}{Diffusion} & Val. & $-2.9_{\pm 0.3}$ & $-3.4_{\pm 0.2}$ & $-2.7_{\pm 0.3}$ & $-3.3_{\pm 0.3}$  \\
                                & Test & $1.0_{\pm 0.4}$ & $0.9_{\pm 0.8}$ & $-2.9_{\pm 0.2}$ & $-3.5_{\pm 0.1}$ \\ \hline
    \end{tabular}
    \caption{Comparison of the log-mse of different hybrid modelling strategies in validation and OOD test settings. \textit{Except on RLC, AHMs always outperform the corresponding hybrid learning models on the test sets. Good performance on the validation set are conserved with augmentation.}}
    \label{tab:synth_log_mse}
\end{table}
\subsection{Distribution shift visualization}
\begin{figure*}
    \centering
    \includegraphics[width=.98\textwidth]{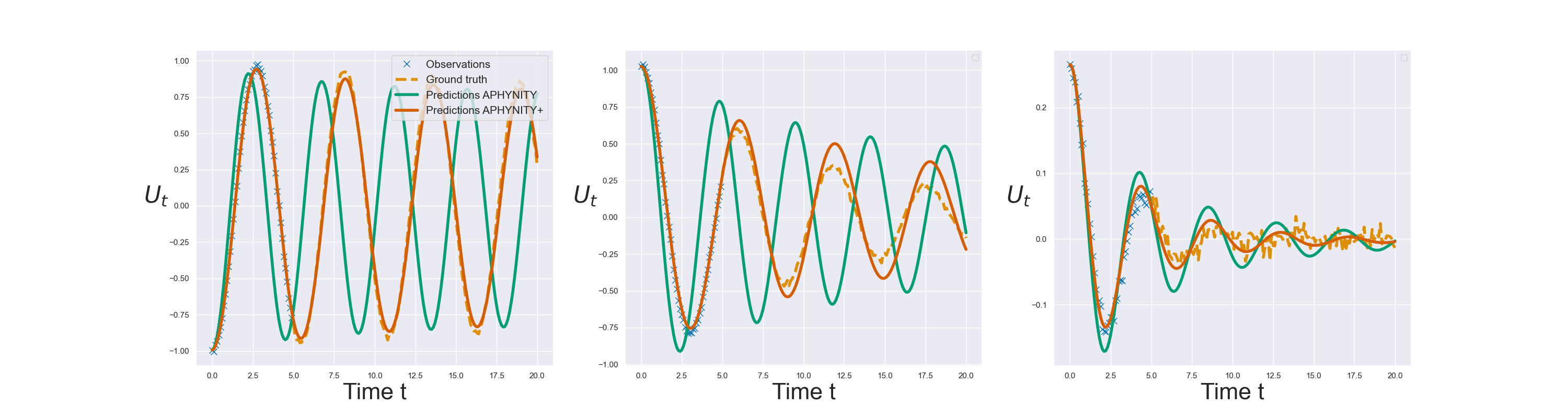}
    \vspace{-1em}
    \caption{Comparison of the predictions made by APHYNITY and APHYNITY+ on the damped pendulum problem for 3 diverse test examples. It is important to mention that the support of the test distribution is disjoint from the training support. \textit{We clearly observe the beneficial effect of augmentation which lead to more accurate predictions.}}
    \label{fig:ood_pendulum}
\end{figure*}

\begin{figure*}
    \centering
    \includegraphics[width=.98\textwidth]{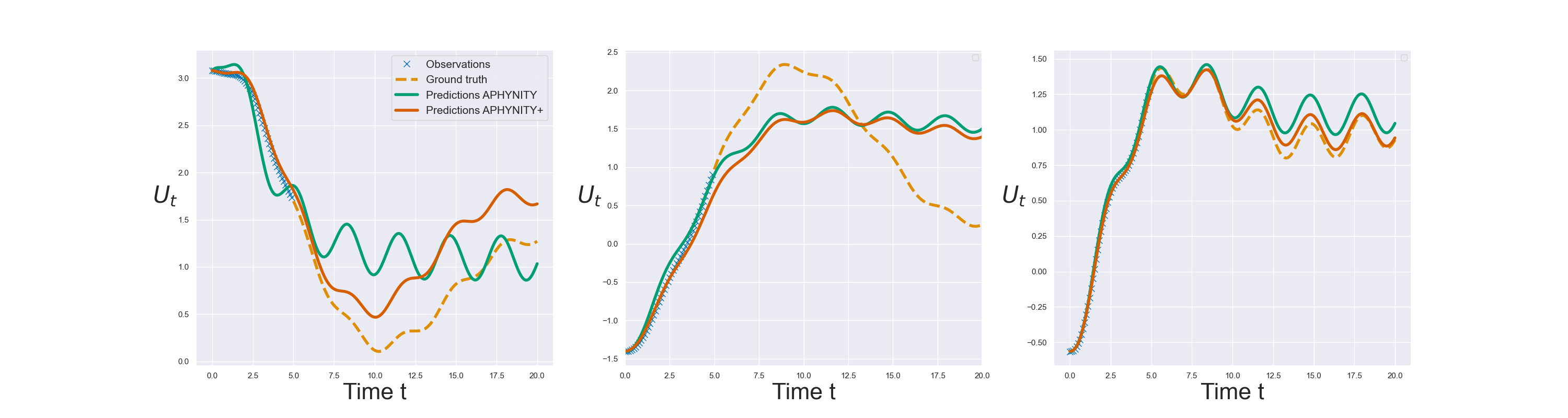}
    \vspace{-1em}
    \caption{Comparison of the predictions made by APHYNITY and APHYNITY+ on the RLC series problem for 3 diverse test examples. It is important to mention that the support of the test distribution is disjoint from the training support. \textit{We can perceive the beneficial effect of augmentation which lead to more accurate predictions in some cases. However both models are inaccurate. This indicates that the RLC series parameters are not easily identifiable, hence the generative model is not exact and augmentation is not as useful as for the diffusion and the pendulum.}}
    \label{fig:ood_RLC}
\end{figure*}

Similar to \figref{fig:diffusion_shift}, \figref{fig:ood_pendulum} and \figref{fig:ood_RLC} showcase the behaviour of APHYNITY and APHYNITY+ for OOD test samples.

\subsection{On the effect of out of expertise shift}
\begin{figure*}[h]
    \centering
    \includegraphics[width=.98\textwidth]{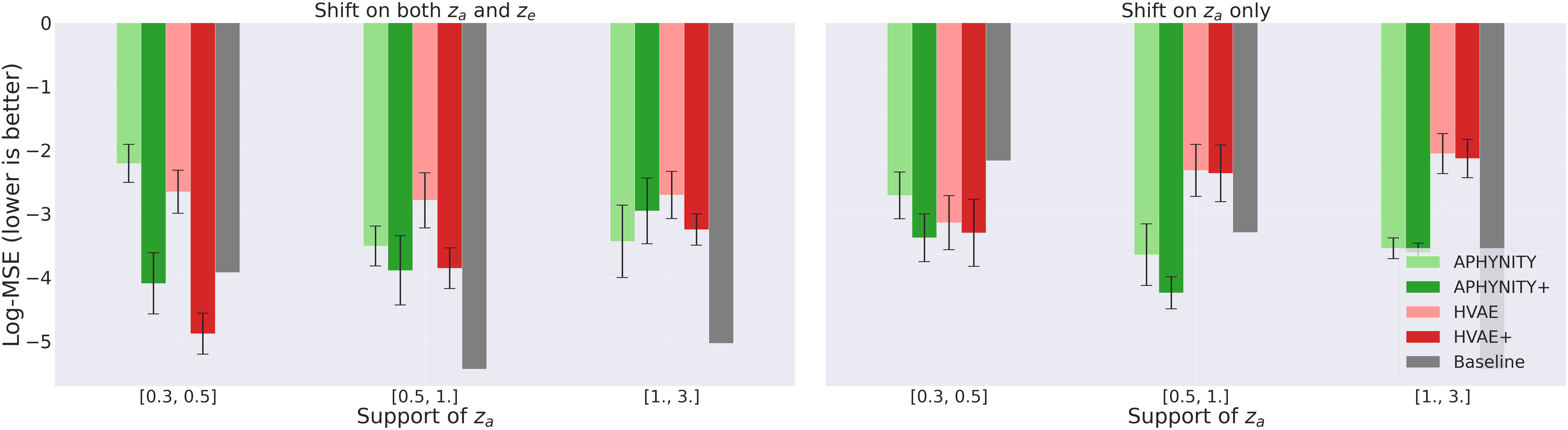}
    \caption{\textbf{Damped pendulum.} Effect of a distribution shift on the latent variable $z_a$ of the interaction model. When the shift of $z_a$ is reasonable (less than $1$), the augmented models outperforms standard models even when the shift is only on $z_a$.}
    \label{fig:za_ood_pendulum}
\end{figure*}

\begin{figure*}[h]
    \centering
    \includegraphics[width=.98\textwidth]{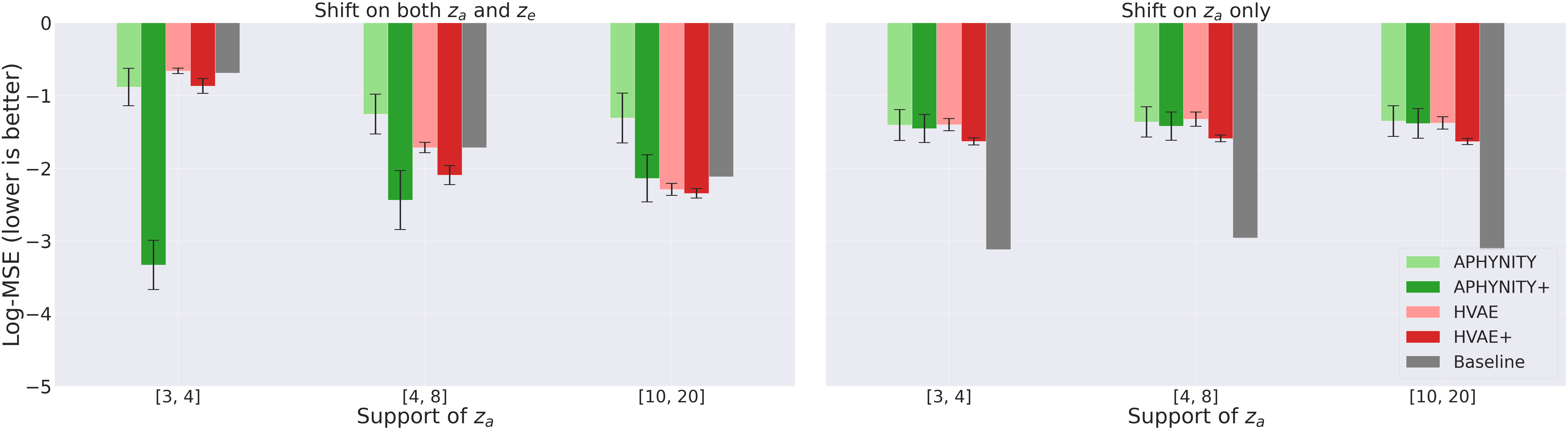}
    \caption{\textbf{RLC series.} Effect of a distribution shift on the latent variable $z_a$ of the interaction model. We observe that augmentation is always beneficial, even when the shift is only on $z_a$. As the dynamics of the RLC series systems depends on the values of all $3$ parameters $R, L, C$, we observe that some distribution shift can even lead to improved performance for the augmented models as for APHYNITY+ when $R \in \left[3, 4\right]$}
    \label{fig:za_ood_RLC}
\end{figure*}

\begin{figure*}[h]
    \centering
    \includegraphics[width=.98\textwidth]{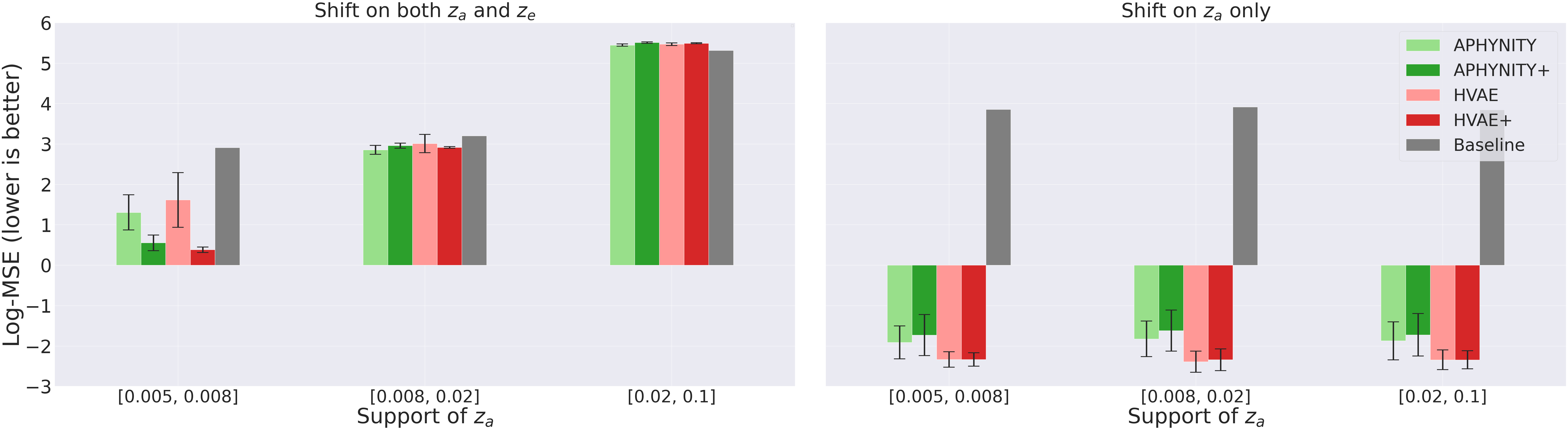}
    \caption{\textbf{2D diffusion reaction.} Effect of a distribution shift on the latent variable $z_a$ of the interaction model. When the shift of $z_a$ is reasonable ($k < 0.008$), the augmented models outperforms standard hybrid learning even when the shift is only on $z_a$.}
    \label{fig:za_ood_diffusion}
\end{figure*}

The additional results in \figref{fig:za_ood_pendulum}, \figref{fig:za_ood_RLC} and \figref{fig:za_ood_diffusion} demonstrate that our augmentations is mostly always beneficial. Although the benefit of augmentation decreases with the gap between the support of the distributions of $z_a$ and train and test times, it still performs either better or on par with non-augmented hybrid learning models.

\subsection{Applying augmentation at multiple stage of the learning} \label{app:add_results_double_pendulum}
\begin{figure}[H]
    \centering
    \includegraphics[width=1.\textwidth]{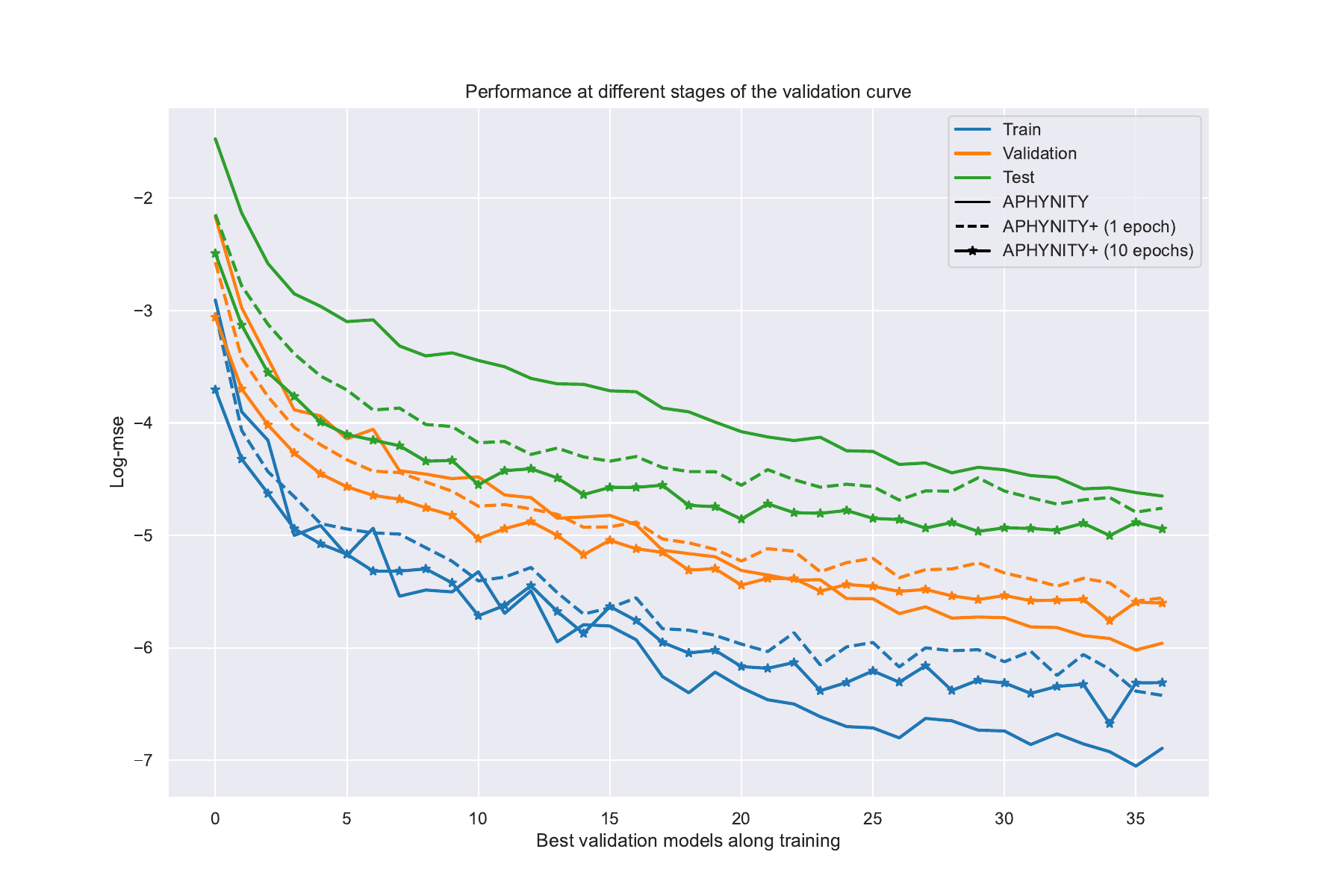}
    \caption{The performance of APHYNITY and APHYNITY+ along training. Plain lines represent the evolution of training, validation and test log-mse of APHYNITY on the double pendulum along training. Dashed lines and stared lines respectively depicts the performance when the corresponding model finetuned respectively for one and ten epochs. \textit{Expert augmentation reduces the gap between the train, validation and test log-mses at any stage of the training. In particular it tends to slightly decrease the test performance and to improve the test performance, as expected from the augmentation strategy.}}
    \label{fig:augmentation_along_training}
\end{figure}

In order to check the robustness of applying the expert augmentation in practical settings, for the quality of the interaction model is unknown, we study the improvement brought by applying augmentation at the different stage of the training. \figref{fig:augmentation_along_training} shows that our augmentation robustly improves the test performance, even when the finetuning is performed on models that are not trained until convergence. We also observe that, as the model gets better trained, it takes more finetuning iterations to maintain the same validation performance. It is why in the results presented in the main text we used 100 epochs of finetuning in order to get the best improvements from the expert augmentation.

\section{Optimal Bayes predictor}\label{app:proof_opt_bayes}
We define the optimal Bayes predictor from a family of approximators $\mathcal{F}$ of a quantity $y$ given an input $x$ as the posterior distribution $q(y|x)$ that has a minimal error on the true posterior $p(x|y)$. We consider the Kullback-Leibler divergence as the objective and formally define the optimal Bayes predictor $p_B(y|x) \in \mathcal{F}$ as 
$$p_B(y|x) = \arg\min_{q \in \mathcal{F}} \mathbb{E}_{p(y|x)}\left[ \log{\frac{p(y|x)}{q(y|x)}} \right].$$

We assume the model in \figref{fig:gen_bnet} and are aiming for the Bayes optimal predictor of $y$ given an input $x$ and a pair of observations $p(x_o, y_o)$ taken from the same system; that is for $y\sim p(Y|x, z_a, z_e)$ and $y_o \sim p(Y|x_o, z_a, z_e)$ coming from the same system's parameters $z_a$ and $z_e$ but potentially different perturbations $x$ (and $x_o)$. We consider a class of predictors expressed as an expectation of a universal conditional distribution $q(z_a, z_e| (x_o, y_o))$ over a universal conditional distribution $q(y|x, z_a, z_e)$. Any elements in $\mathcal{F}$ is defined as $\mathbb{E}_{q(z_a, z_e| (x_o, y_o)}\left[ q(y|x, z_a, z_e) \right]$ where $q(z_a, z_e| (x_o, y_o))$ and $q(y|x, z_a, z_e)$ can be any continuous conditional distribution. We also consider that all quantities discussed belong to $\mathbb{R}^k$ for $k \in \mathbb{N}$ and have continuous support. 
We aim to prove that $q(z_a, z_e| (x_o, y_o)) = p(z_a, z_e| (x_o, y_o))$ and $q(y|x, z_a, z_e) = p(y|x, z_a, z_e)$ constitutes the Bayes optimal predictor of $y$ given $(x, x_o, y_o)$ in $\mathcal{F}$. Indeed, we have:
\begin{align}
    \mathbb{E}_{p(z_a, z_e| (x_o, y_o))}\left[ p(y|x, z_a, z_e) \right] &= \int p(z_a, z_e| (x_o, y_o)) p(y|x, z_a, z_e) dz_a dz_e \\
    &= \int p(z_a, z_e| (x_o, y_o)) p(y|x, z_a, z_e,  (x_o, y_o)) dz_a dz_e \\
    &= \int p(y, z_a, z_e|x,  (x_o, y_o))dz_a dz_e  \\
    &= p(y|x,  (x_o, y_o)),
\end{align}
and thus
\begin{align}
    \mathbb{KL}\left[p(y|x, x_o, y_o) || \mathbb{E}_{p(z_a, z_e| (x_o, y_o))}\left[ p(y|x, z_a, z_e) \right] \right] = \mathbb{KL}\left[p(y|x, x_o, y_o) || p(y|x,  (x_o, y_o)) \right]  = 0
\end{align}
which is the minimal value of the objective function and is thus optimal.

% \bibliography{main}
% \bibliographystyle{tmlr}

% \appendix
\end{document}